\pdfoutput=1
\documentclass[11pt]{article}

\usepackage[final]{acl}

\usepackage{times}
\usepackage{latexsym}

\usepackage{soul,url,amsmath,amsthm,amssymb,epsfig,epstopdf,multirow,bm,enumitem,url,bbm,colortbl,booktabs,amsfonts,colortbl,tgcursor,makecell,xcolor}

\soulregister{\cite}7 % 注册\cite命令
\soulregister{\citep}7 % 注册\citep命令
\soulregister{\citet}7 % 注册\citet命令
\soulregister{\ref}7 % 注册\ref命令
\soulregister{\pageref}7 % 注册\pageref命令
\graphicspath{ {Images/} }
\usepackage[linesnumbered, ruled]{algorithm2e}
\SetKwRepeat{Do}{do}{while}%

\usepackage[T1]{fontenc}
\usepackage[utf8]{inputenc}

\usepackage{microtype}

\usepackage{inconsolata}

\title{Multimodal Clickbait Detection by De-confounding Biases Using Causal Representation Inference}

\author{
    \hspace{-1.5mm}Jianxing Yu\hspace{-0.3mm}\thanks{~~Corresponding author.}$\hspace{1.2mm}^\dag$\hspace{-1mm}, Shiqi Wang\hspace{-0.3mm}\thanks{~~These authors have contributed equally to this work.}\hspace{0.6mm}, Han Yin\hspace{-0.3mm}$^\dag$\hspace{-1mm}, Zhenlong Sun, Ruobing Xie, Bo Zhang, Yanghui Rao \\
  School of Artificial Intelligence, Sun Yat-sen University, Zhuhai, 519082, China \\
  International Campus, Zhejiang University, Haining, 314400, China \\
  \normalsize{Key Laboratory of Sustainable Tourism Smart Assessment Technology, Ministry of Culture and Tourism}\\
  \normalsize{WeChat Search Application Department, Tencent, Beijing, 100080}\\
  Pazhou Lab, Guangzhou, 510330, China \\
  \normalsize{\{yujx26, wangshq25, raoyangh\}@mail.sysu.edu.cn, 
  han.22@intl.zju.edu.cn} \\
  \normalsize{richardsun@tencent.com, xrbsnowing@163.com, nevinzhang@tencent.com}
}

\begin{document}
\maketitle
\begin{abstract}
This paper focuses on detecting clickbait posts on the Web. These posts often use eye-catching disinformation in mixed modalities to mislead users to click for profit. That affects the user experience and thus would be blocked by content provider. To escape detection, malicious creators use tricks to add some irrelevant non-bait content into bait posts, dressing them up as legal to fool the detector. This content often has biased relations with non-bait labels, yet traditional detectors tend to make predictions based on simple co-occurrence rather than grasping inherent factors that lead to malicious behavior. This spurious bias would easily cause misjudgments. To address this problem, we propose a new debiased method based on causal inference. We first employ a set of features in multiple modalities to characterize the posts. Considering these features are often mixed up with unknown biases, we then disentangle three kinds of latent factors from them, including the invariant factor that indicates intrinsic bait intention; the causal factor which reflects deceptive patterns in a certain scenario, and non-causal noise. By eliminating the noise that causes bias, we can use invariant and causal factors to build a robust model with good generalization ability. Experiments on three popular datasets show the effectiveness of our approach.
% \hl{The noise factor contains insignificant content which co-occurs frequently with bait/non-bait labels, and would induce detectors to make biased decisions. By using invariant and causal factors solely, we de-confound biased correlations to build a robust model with good generalization ability.} Experiments on three popular datasets show the effectiveness of our approach.

%One is the invariant discriminant factor that can make consistent predictions in most scenarios. The scenario is used to describe the bait patterns and writing styles of posts in certain periods, types, and creators. Each is estimated by alternating optimization of disentanglement and prediction. In addition, we de-confound the causal factor in a specific scenario and non-causal factor by causal inference with contrastive constraints. 

\end{abstract}

\section{Introduction}

With the rapid development of social media, people share views and advertise products by posting content on platforms~\cite{liao2021investigating} like WeChat, Twitter, Instagram, etc. To increase viewership and obtain more advertising revenue, unscrupulous creators misuse these platforms to publish deceptive, poor-quality posts. Such posts are often described with a catchy thumbnail or a sensational headline. For example, the posts have a sexy thumbnail, with curiosity-inciting phrases like ``\emph{You Won't Believe}'', and ``\emph{X Reasons Why}'' in their headlines. That would bait readers to click on the linked articles. However, these articles would be unrelated to the thumbnails and headlines of the posts. Moreover, they are often full of hoaxes, rumors, and fake news, which not only degrade users' experience but also affect the credibility of the platforms~\cite{zhu2023clickbait}. Thus, the detection of such clickbait posts has great commercial value for social media. 

%For example, a bait post would have lots of retweets in a short period, but the reading time is short due to its poor quality. 
Due to the large volume of emerging posts, a manual review of the clickbait is infeasible. Machine detection has become a hot topic~\cite{comito2023multimodal}. The detective sources can be summarized into two categories~\cite{yadav2023comparative}. The first is based on social behavior. Since bait posts are required to spread in social networks to expand their influence, typical propagation characteristics can be observed. It can be detected based on social metadata like comments, the number of views, shares, likes, etc~\cite{agarwal2023understanding}. That requires a rich collection of user feedback for judgment, but this feedback is often delayed or some users do not even share it. As a result, malicious posts can only be found after they have been widely spread, which is too late for online applications. Another direction is to analyze the post contents. The bait posts often have certain linguistic characteristics in terms of deceptive words, syntax, subjectivity, writing style~\cite{zhou2020fake}, and even punctuation~\cite{coste2021advances}. Besides, there may be abnormal relations between their sub-parts in various modalities, such as a rumor article text with visual thumbnails of an irrelevant actor to attract clicks. Early works design a set of rules to detect these features and relations, but the rules are hand-crafted and non-scalable. Current mainstream methods turn to the neural model for improving scalability. They seek correlations between the post content and labels to make predictions.
\begin{figure}[h]
\centering
\includegraphics[width=\linewidth]{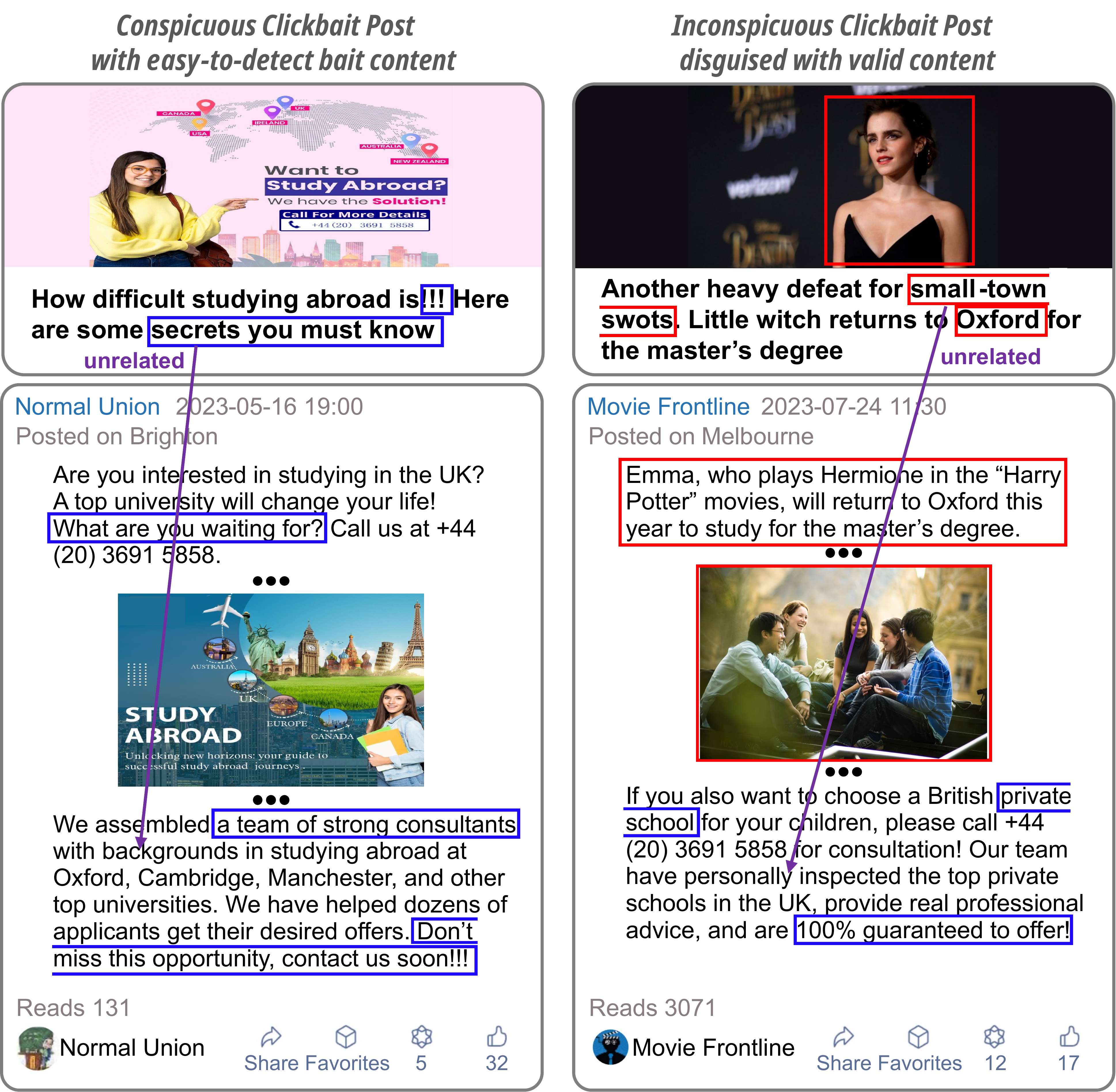}
\caption{Clickbait samples. The purple arrow indicates inconsistencies between the headline and its linked article. The simple clickbait post contains conspicuous bait-indicative words or advertising content (marked with \textcolor{blue}{blue boxes}) which is easily detected. The complex one disguises the bait content with some valid content (in \textcolor{red}{red boxes}) and makes it look inconspicuous, thus deceiving and escaping the detector.}
\label{fig:sampleq}
\end{figure}

However, to escape detection, malicious creators would use some unrelated content to disguise a bait post as a valid one. As shown in Fig.(\ref{fig:sampleq}), in the left post, some of the conspicuous bait thumbnail and text are replaced with unharmful content, making it look like a legal one on the right. Such content may co-occur with non-bait label usually, creating spurious correlations, i.e., the unstable and confounding patterns in the feature space~\cite{dogra2022comparative}. That would easily cheat the model to fail to discover the hidden bait. Respectively, some valid posts may be misjudged as bait simply because they contain content that co-occurs with a bait label. Such biased correlations lead to high misjudgments. The bias is inadequate to differentiate by a naive encoder, which is designed for generic content understanding tasks such as text classification. The encoder focuses on general embedded representations rather than key factors that cause fraudulent behaviors~\cite{mukherjee2022did}. These behaviors are not static, their topics and types may change over time. The writing styles would also vary for different authors, but such disguise tricks commonly follow some styles and patterns in a particular scenario. On the other hand, the data-driven neural model relies on scarce labeled data heavily. To derive a robust model, we have to collect data on all scenarios for supervised training. Such uniform data is either unavailable or costly to acquire. As a result, the statistics on limited samples are not significant, which is easy to train the model to remember some inessential spurious correlations. That would harm the model's robustness. A simple solution is to customize a representation that perceives posts' quality, but that would face tedious hand-crafted engineering. 

Motivated by above observations, we propose a new debiased approach to detect clickbait posts. In detail, we first represent the given posts based on a set of multimodal features, including textual and visual features, linguistic and cross-modal features, as well as features of the creators' profiles. Considering these mixed representations contain unknown biases, we resort to causal representation learning, which is good at eliminating non-causal spurious noise. The representations are disentangled into three key latent factors, including (1) invariant factor that indicates the inherent bait's intention and post quality; (2) causal factor in a specific scenario; and (3) non-causal noise factor. The noise like unrelated words and topics is defined under a specific scenario, and each is unique. The invariant factor is obtained by invariant risk minimization, scenarios are estimated from the data, and the causal factor is learned by contrastive learning. By removing the noise and using the remaining invariant and causal factors, we can build a robust clickbait detector. It can combat spurious bias and generalize well on newly formed bait subspecies. To facilitate training, we further develop a data augmentation technique to alleviate the labeled data scarcity problem. Extensive experiments on three real-world datasets show the effectiveness of our approach. 

The main contributions of this paper include,
\begin{itemize}
  \item
  We reveal the issue of bait subspecies evolution via disguise and point out the challenges of the resulting spurious bias in the field of multimodal clickbait detection, which is new.
  \item
  We propose a new debiased model from a view of causal inference. It explores a prior causal structure to elicit latent key factors that reflect posts' quality. That can alleviate spurious bias and achieve better generalization ability. 
  \item
  We conduct extensive experiments to fully evaluate the effectiveness of our method.
  \end{itemize}

%The rest of this paper is organized as follows. Section 2 elaborates on our proposed causal model. Afterward, Section 3 presents the experimental results. Section 4 reviews the related works. Finally, Section 5 concludes this paper with future works.

\begin{figure*}[h]
\centering
\includegraphics[width=\linewidth]{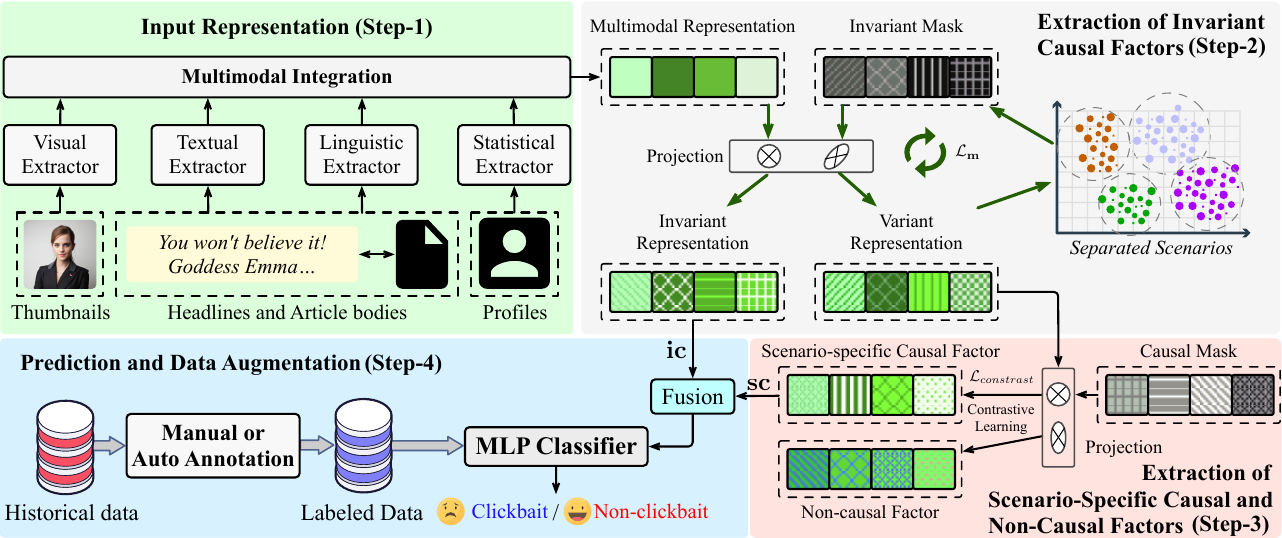}
\caption{The overview framework of our causal clickbait detector.}
\label{fig:overview}
\end{figure*}

\section{Approach}\label{sec:appoverview}

Fig.(\ref{fig:overview}) shows our framework with four steps. We first extract multimodal features from the posts. By causal inference, we then disentangle the invariant factors from them, and separate the remaining parts into the causal factor and non-causal factor. Finally, we make the prediction based on the invariant and causal factors. Next, we define some notations and elaborate on each component of our approach.

% Fig.(\ref{fig:overview}) shows our framework with three components, including extraction of the multimodal features on posts, disentanglement of key causal factors from these features, and making predictions based on such causal representations with affordable training cost. Next, we define some notations and elaborate on each component of our approach.

\subsection{Notations and Problem Formulation}
\begin{figure}[h]
\centering
\includegraphics[width=2.7in]{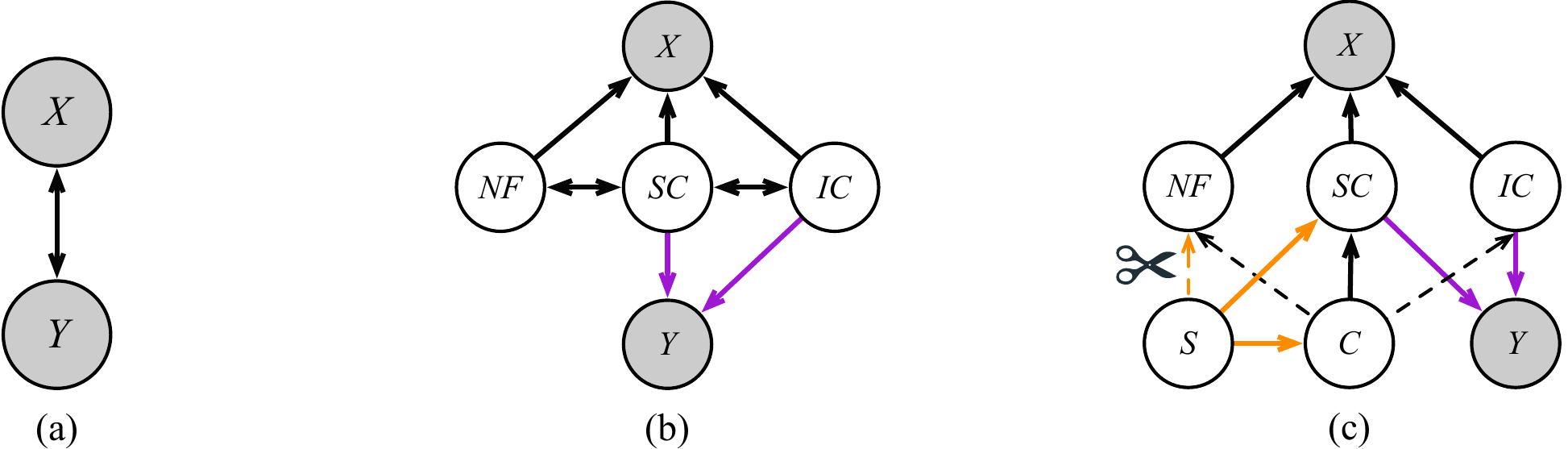}
\caption{Causal structure for de-confounding biases. Gray and white nodes represent observable and unobserved variables, respectively; the bidirectional and unidirectional arrows denote correlations and causalities, respectively; purple arrows are key causalities determining result $Y$; and orange arrows refer to scenario effects.}
\label{fig:causalstructure}
\end{figure}

\textbf{Clickbait detection}. Given a post $x_i$, our task aims to learn a model $\mathcal{F}(y_i | x_i, \Theta_\mathcal{F})$ to predict whether it is clickbait ($y_i=1$), where $\Theta_\mathcal{F}$ is the model's parameter set, $x_i$ denotes the post contents, including the textual headline, visual thumbnail and their linked mixed-modal article. These sub-parts have some deceptive characteristics and relations, such as containing malicious rhetoric like exaggeration, eroticism, bluffing, weirdness, and distortion; the headline or thumbnail is seductive but unrelated to the article. In addition, the creators constantly yield new bait subspecies to avoid being detected and seized. Thus, our target is to find an optimized $\hat{\Theta}_\mathcal{F}$ by $\arg \min_{\hat{\Theta}_\mathcal{F}} \mathcal{L}(\mathcal{\hat{F}}(y_i|x_i, \hat{\Theta}_{\mathcal{F}})| \mathcal{D}^{tr})$ which can be generalized well to the test set $\mathcal{D}^{te}$ with a lowest cost $\mathcal{L}(\mathcal{D}^{te})$, where $\mathcal{D}^{tr}$ is the training set, $\mathcal{L}(\cdot)$ denotes a cross-entropy classification loss.\\
\textbf{Eliminating spurious correlations}. As displayed in Fig.\ref{fig:causalstructure}.(a), traditional methods often make predictions based on the co-occurrence between post features $X$ and labels $Y$. Ideally, $X$ is expected to reflect the hidden malicious intention of the post. However, it comes from generic encoders without considering the fraud behavior. That inevitably introduces some false correlations, such as wrongly linking some trivial but irrelevant terms or images with non-bait label. Besides, some bait modes are stable while some writing styles would change in various scenarios, such as periods, types, and creators. It is easy to make erroneous predictions without grasping this distinction. A simple solution is to dissociate $X$ into several factors, i.e., an invariant factor ($IC$) that reflects malicious intention, a causal factor for a certain scenario ($SC$), and non-causal noise ($NF$); and then eliminate the effect of noise $NF$ on $Y$, as Fig.\ref{fig:causalstructure}.(b). However, without necessary constraints, the spurious correlations in $NF$ may permeate via $IC$ and $SC$ to harm the prediction of $Y$. To prevent this effect, we introduce a causal structure to regularize these latent variables, as Fig.\ref{fig:causalstructure}.(c). We first use a confounder $C$ to capture mixed relations of three factors under the condition of a certain scenario $S$. We then isolate the invariant $IC$ by blocking the influence of $C$ on $IC$. The bias in $C$ cannot affect $Y$ through $IC$. To fully put away the noise $NF$, we cut the backdoor paths of $C$ and $S$ on $NF$. The $S$ and $C$ can only impact $Y$ via $SC$. In this way, we can independently elicit salient unbiased factors without mutual influence, which can generalize well to new bait subspecies.

%(a)传统方法捕捉输入特征X与标签Y的共现关系来检测标题党. (b) 本文将X分解为不变特征，以及情景特定的因果特征。其中SC和IC持有标题党判别的充足因果条件. (c) These features are entangled due to the confounding effects of backdoor confounder and domain selector 。通过阻断后门路径，我们的模型得以在预测时消除混淆的干扰。Causal structure of clickbait detection. White nodes represent latent features; gray nodes are observable variables; bidirectional arrows denote correlations; unidirectional arrows represent causal conditions; red arrows are causal conditions critical to prediction; and green arrows refer to scenario effects.

\subsection{Multimodal Feature Extraction}\label{sec:inputrepresentation}

To fully capture the characteristics of the posts, we extract five kinds of features in multiple modalities. More details are exhibited in the Appendix~\ref{sec:linguistic}.\\
\textbf{(1) Visual Features}: We encode each post image (e.g. thumbnail and article figure) by transformer-based \emph{Swin-T}~\cite{liu2021swin}. To grasp the fine-grained features in the image, we conduct facial recognition and object detection by \emph{DNN} and \emph{RetinaNet} models~\cite{lin2017focal}, respectively. \\
\textbf{(2) Textual Features}: For the text in the post, such as headline, description, and associated article, we first remove hashtag, mention, punctuation, and URL. We then tokenize each and embed it by BERT$_{base}$ pre-trained model~\cite{Devlin2018}. Since clickbait posts often contain seductive text in the thumbnails, we utilize the \emph{OCR} technique~\cite{li2023trocr} to extract the text and encode it.\\
\textbf{(3) Cross-modal Features}: The clickbait posts often contain inconsistencies, such as using inviting thumbnails to lure clicking the unrelated article. Verifying this cross-modal unmatch between the post's sub-parts can help prediction. Due to the heterogeneous gap, it is hard to use this inter-modal complementary benefit directly. To tackle this issue, we employ a \emph{CT} Transformer~\cite{lu2019vilbert}, which is good at capturing multi-modal relations. \\
\textbf{(4) Linguistic Features}: To capture the bait patterns and writing styles, we extract six kinds of features based on thumbnail, headline, and article body, including article body-thumbnail disparity, article body-headline disparity, thumbnail-headline disparity, sentiment of headline, lexical analysis of headline and baitiness analysis of headline.\\
\textbf{(5) Profile Features}: The posts' quality often depends on the creators, i.e., bad creators tend to write malicious posts. To characterize the quality of each creator $u_j$, we extract features based on its profiles, involving the register age, self-description and screen name of $u_j$, number of followers for $u_j$, number of users that $u_j$ is following, number of created posts of $u_j$, time elapsed after $u_j$'s first post, whether the $u_j$ account is verified or not, whether $u_j$ allows the geo-spatial positioning, time difference between the source post time and $u_j$'s share time, and length of retweet path between $u_j$ and a source post. The length is 1 if $u_j$ retweets the post.\\
By concatenating all these features, we can obtain a $d$-dimensional vector $\mathbf{x}_{i}$ as a post's representation.

%%有两类特征，不变和可变，先用IRM分离出来；然后用对比学习把可变特征分离出来
%%可变特征也分两类，存在因果和不存在因果，拿对抗学习，通过屏蔽机制分离出来
\subsection{Disentanglement of Invariant Factors}\label{sec:invariant}

The generic $\mathbf{x}_{i}$ may contain spurious correlations, which are mostly unreliable across various scenarios. For example, the false correlation between an unrelated actress and the bait label will change in different contexts of posts. In contrast, it is usually stable for the bait behavior which is caused by some common factors, such as underlying language patterns and deceptive habits. By seeking commonalities in various scenarios, we can capture these inherent factors to achieve better robustness.

\textbf{Casual Inference}: We use an invariance mask $\mathbf{m} \in \mathbb{R}^{d}$ to dissociate the invariant characteristics of the original representation $\mathbf{x}_{i}$ as $\mathbf{ic}_{i} = \mathbf{m} \odot \mathbf{x}_{i}$, where $\odot$ is the element-wise product operator. The opposite of $\mathbf{m}$ is the variance mask, which can extract the variant representation $\mathbf{vc}_{i} = (1 - \mathbf{m}) \odot \mathbf{x}_{i}$. Ideally, $\mathbf{ic}_{i}$ should have identical joint distributions with the unbiased variable across various scenarios. We pursue it by optimizing $\mathbf{m}$ with an invariant risk minimization (\emph{IRM})~\cite{arjovsky2019invariant} objective. \emph{IRM} introduces a penalty for the variation of empirical risks in all scenarios. It tries to mask some features and calculate their impact on the results, so as to find out factors that have a stable and significant impact on the results. That encourages the mask to suppress spurious features and emphasize the causally invariant ones. We formulate \emph{IRM} loss as a gradient norm penalty over the empirical risk $\mathcal{L}_{s}$ in each training scenario as Eq.(\ref{formula:opobj}):
\begin {equation}
\begin{split}
\mathcal{L}_{\mathbf{m}} = &\frac{1}{|\mathcal{S}|} {\textstyle \sum\nolimits_{s \in \mathcal{S}} [\mathcal{L}_{s}}(\mathbf{m}, \mathcal{F}_{\mathbf{m}}) +\\
& \alpha ||\triangledown_{\mathcal{F}_{\mathbf{m}}} \mathcal{L}_{s}(\mathbf{m}, \mathcal{F}_{\mathbf{m}})||^2] + \beta ||\mathbf{m}||^2.\\
\end{split}
%问题：\mathop{{\min }}\nolimits_{{\textbf{m}, \mathcal{F}}} \frac{1}{|\mathcal{S}|} {\textstyle \sum\nolimits_{s \in \mathcal{S}} [\mathcal{L}_{s}}(\mathbf{m}, \mathcal{F}) + \alpha ||\triangledown_{\mathcal{F}} \mathcal{L}_{s}(\mathbf{m}, \mathcal{F})||^2] + \beta ||\mathbf{m}||^2\\
\label{formula:opobj}
\end {equation}
where $\alpha$ and $\beta$ are trade-off factors, the second term is the constraint over the variance of scenarios, and the third one is a regularization term. {$\mathcal{L}_{s} = \mathcal{L}(\mathcal{F}_{\mathbf{m}}(y_{i}|x_{i},\Theta_{\mathcal{F}_{\mathbf{m}}})|\mathcal{D}_{s}^{tr})$} is the classified loss on the training subset $\mathcal{D}_{s}^{tr}$ under the specific scenario $s$. This objective can enforce the mask $\mathbf{m}$ to seek stable inherent patterns in the context, instead of learning an average effect of spurious correlations. 

\textbf{Scenario Estimation}: The key to optimizing $\mathbf{m}$ lies in the division of scenarios. We thus utilize scenarios to infer the spurious correlations and determine each with a set of operations. In each scenario, the posts have certain characteristics, such as writing styles, hot topics, and evolving bait patterns. Since the scenario is unprovided, we propose to estimate it from the data with two iterative phases. First, we learn a scenario predictive model to fit the data distribution. We then reassign the samples into appropriate scenario subsets. To initialize this iterative process, we randomly assign training samples to each scenario subset $\mathcal{D}_{s}^{tr}$. By alternatively optimizing the data-fit \emph{IRM} objective and the estimated scenarios, finally we can learn the invariant factors across scenarios. \\
\textit{(1) Scenario Division.} We observe that the key difference among various scenarios lie in their characteristics of spurious correlations. We thus explore such correlations to learn a scenario predictive model $\Phi_{s}$ for each $s \in \mathcal{S}$. In detail, for the $i^{th}$ sample, $\Phi_{s}$ evaluates the likelihood that it belongs to the scenario $s$ based on its variant representation $\mathbf{vc}_{i}$. Parameterized by $\theta_{s}$, the model $\Phi_{s}$ aims to minimize the prediction loss on the training subset $\mathcal{D}_{s}^{tr}$, as Eq.(\ref{formula:scenario_learning}). Relying on this objective, $\Phi_{s}$ can be learned without interference from other scenario data, which can better handle the spurious bias.
\begin {equation}
\mathop{{\min }}\nolimits_{\theta_{s}} \mathcal{L}(\Phi_{s}(\mathbf{vc}_{i}|\theta_{s})|\mathcal{D}_{s}^{tr}).\\
\label{formula:scenario_learning}
\end {equation}
\textit{(2) Samples Reallocation.} Based on the estimation in phase 1, we obtain $|\mathcal{S}|$ scenario models which indicate different types of spurious biases. Next, we reallocate all samples to the appropriate scenarios according to their spurious correlations. In detail, we feed the variant representation $\mathbf{vc}_{i}$ into each scenario model $\Phi_{s}$. Each sample is then assigned to a scenario with the highest likelihood as Eq.(\ref{formula:data_division}).
\begin {equation}
s(i) \leftarrow \mathop{\arg\max}\nolimits_{s \in \mathcal{S}} \Phi_{s}(\mathbf{vc}_{i}|\theta_{s}).\\
\label{formula:data_division}
\end {equation}
By running these two phases until convergence, we obtain stable scenario subsets $\{\mathcal{D}_{s}^{tr}|s \in \mathcal{S}\}$. In turn, we can optimize the invariance mask $\mathbf{m}$. Finally, we learn an optimal $\mathbf{m}$ to derive a universal causal factor that is invariant in most situations.

\subsection{Dissociation of Causal Factor from Noise}

After separating the invariant factor, the remaining part $\mathbf{vc}_{i}$ is a mixture of causality and noise in a specific scenario. We thus further elicit the valuable scenario-specific causal factor $\mathbf{sc}_{i}$ from $\mathbf{vc}_{i}$. We first project it into a latent embedding space, as $\xi(\mathbf{vc}_{i})$. $\xi(\cdot)$ is a network with a multi-layer perceptron, which is parameterized by $\theta_{\xi}$. When designing $\xi$, we do not inject any scenario category information. That allows us to adapt to some new scenarios without refactoring $\xi$ or relearning the parameters. We then output a causal perception mask $\bm{\gamma} = Gumbel-SoftMax(\xi(\mathbf{vc}_{i}),kd)$ using the \emph{Gumbel-SoftMax} technique. The mask $\bm{\gamma}$ sets the values of some dimension to $0$ and retains $kd$ effective dimensions. By using the mask $\bm{\gamma}$, we can extract the causal representation as $\mathbf{sc}_{i} = \bm{\gamma} \odot \mathbf{vc}_{i}$.
%In particular, we project $\mathbf{vc}_{i}$ into a latent embedding space, as $\xi(\mathbf{vc}_{i})$, where $\xi(\cdot)$ is a network with a multi-layer perceptron. It can output a causal perception mask $\bm{\gamma} = Gumbel-SoftMax(\xi(\mathbf{vc}_{i}),kd)$ using the \emph{Gumbel-SoftMax} technique. The mask $\bm{\gamma}$ sets the values of some dimension to $0$, and retains $kd$ effective dimensions. When designing $\xi$, we do not inject any scenario category information. That allows us to adapt to some new scenarios without refactoring $\xi$ or relearning the parameters. By using the mask $\bm{\gamma}$, we are able to extract the scenario-specific causal representation as $\mathbf{sc}_{i} = \mathbf{vc}_{i} \cdot \bm{\gamma}$.

To facilitate the separation of causal factors from noises, we introduce contrastive constraints based on the causal interventions. After the causal vector $\mathbf{sc}_{i}$ is extracted from $\mathbf{vc}_{i}$, we collect the remaining non-causal one $\mathbf{nf}_{i} = (1-\bm{\gamma}) \odot \mathbf{vc}_{i} $. $\mathbf{nf}_{i}$ can be used as a contrastive learning signal with the loss as {Eq.(\ref{formula:contrastive_loss})}. For a clickbait sample ($y=1$), its non-causal feature $\mathbf{nf}_{i}$ exactly blocks out the identifiable clickbait characteristics. When replacing the scenario-specific causal representation $\mathbf{sc}_{i}$ with $\mathbf{nf}_{i}$, the prediction is supposed to be the opposite, i.e., as non-clickbait. Accordingly, for the non-clickbait sample  ($y=0$), since $\mathbf{sc}_{i}$ does not indicate any clickbait characteristics, using $\mathbf{nf}_{i}$ for detection would not change the prediction.
\begin {equation}
{\mathcal{L}_{contrastive} = \mathcal{L}(\mathcal{F}_{\xi}(0|\mathbf{nf}_{i}, \Theta_{\mathcal{F}_{\xi}})|\mathcal{D}^{tr})}. 
\label{formula:contrastive_loss}
\end {equation}
%As shown in Eq.(\ref{formula:contrastive_loss}), the contrastive loss consists of two parts, that is, $\mathcal{L}_{1} $ to ensure the causal sufficiency, and $\mathcal{L}_{2}$ to prevent the degenerate solutions. These losses are computed respectively on the training subsets labeled as clickbait ($y=1$) and non-clickbait ($y=0$), and are balanced by the weight factor $\sigma$. In particular, for the sample $i$ in $\mathcal{D}_{1}^{tr}$, its non-causal feature $\mathbf{nf}_{i}$ exactly blocks out the identifiable clickbait characteristics. Switching to $\mathbf{nf}_{i}$ for detection is supposed to subvert the original prediction result. Therefore, we evaluate the disparity between the non-clickbait label and the outputs of the contrastive samples as $\mathcal{L}_{1}$, and expect to reduce the disparity. Similarly, for the sample $j$ in $\mathcal{D}_{0}^{tr}$, scenario-specific causal representation $\mathbf{sc}_{j}$ does not indicate any clickbait characteristics, so using $\mathbf{nf}_{j}$ as the basis for detection would not change the prediction result. Consequently, the disparity computed in $\mathcal{L}_{2}$ should also be as small as possible. Note%
%\begin {equation}
%\begin{array}{l}
%{\mathcal{L}_{contrastive} = \sigma \mathcal{L}_{1} + (1-\sigma) \mathcal{L}_{2}}, \\
%{\mathcal{L}_{1} = \mathcal{L}(\mathcal{F}(0|\mathbf{nf}_{i}, \theta_{\mathcal{F}})|\mathcal{D}_{y=1}^{tr})}, \\
%{\mathcal{L}_{2} = \mathcal{L}(\mathcal{F}(0|\mathbf{nf}_{j}, \theta_{\mathcal{F}})|\mathcal{D}_{y=0}^{tr})}.
%\end{array}
%\label{formula:contrastive_loss}
%\end {equation}

\subsection{Prediction and Data Augmentation}\label{sec:debiasedclassifier}

By repeating the running flow (i.e., vector masking $\to$ scenario division $\to$ mask learning) for $T$ times till converged, we can obtain the key causal representations, namely, the invariant causal factor and scenario-specific causal factor. We concatenate these two vectors to train a \emph{Multilayer Perceptron} classifier, with an objective as Eq.(\ref{formula:classifierobj}). 
\begin {equation}
\begin{array}{l}
\mathop{\arg\min}\nolimits_{\hat{\Theta}_\mathcal{F}} \mathcal{L}(\hat{\mathcal{F}}(y_{i}|[\mathbf{ic}_{i};\mathbf{sc}_{i}], \hat{\Theta}_\mathcal{F})|\mathcal{D}^{tr}).\\
\end{array}
\label{formula:classifierobj}
\end {equation}\\
To better train the model, we further employ data augmentation to collect pseudo-labeled data. Since clickbait posts need to be widely spread to gain benefits, social behavior is often an effective clue. We thus design heuristic rules based on social metadata, such as share frequency, and viewing time. This metadata may be lacking in new cases, but it is sufficient in historical data. That can provide abundant clickbait cases as training data to reduce labeled costs. In this way, our model can work well even if only limited data is provided in some applications. More details are shown in Appendix~\ref{sec:augmentationdetail}.

%%%%%%%%%%%%%%%%% 开始实验部分 %%%%%%%%%%%%%%%%%
\section{Evaluations}\label{sec:evaluations}

We fully conducted experiments with qualitative and quantitative analyses to evaluate our approach.

\subsection{Data and Experimental Settings}
\label{sec:expsetting}
We performed evaluations on three popular real-world datasets, including \emph{CLDInst}~\cite{DVN/DEZMRA_2018}, \emph{Clickbait17}~\cite{potthast2018crowdsourcing} and \emph{FakeNewsNet}~\cite{shu2018fakenewsnet}. By crowd-sourcing, these datasets were split as bait/non-bait sets with the size of 4k/3k, 9k/29k, and 5k/17k posts, respectively. For each sample, we crawled the original post from its \textit{URL} to collect multimodal data, such as the thumbnail and creator's profile.

 $\bullet$ \emph{CLDInst} covers 7,769 fashion-related posts crawled from \textit{Instagram}. They are judged by annotators employed from crowdsourcing websites. A total of 4,260 posts are tagged as clickbait.

 $\bullet$ \emph{Clickbait17} contains 38,517 Twitter posts as well as their linked articles from 27 major US news publishers. To avoid bias, a maximum of 10 posts per day was sampled for each publisher, with 9,276 posts tagging as clickbait.

 $\bullet$ \emph{FakeNewsNet} is a large-scale multimodal news dataset that contains over 23k articles with tagged fake/real labels from the websites of \emph{PolitiFact} and \emph{GossipCop}. It has a rich social context, with 432 fake and 624 real samples from \emph{PolitiFact}, as well as 5,323 fake and 16,817 real cases from \emph{GossipCop}. Similar to bait posts, fake samples often suffer from issues like irrelevance, inconsistency, etc. This dataset can be used to evaluate the model's ability to recognize bait-like low-quality content.
 
Each dataset was split into train/validation/test sets. We tuned the model on a validation set and reported results on the test set. In addition, we further evaluated the value of data augmentation technique in Appendix~\ref{sec:augmentationdetail}. We employed four typical metrics in the field of classification for evaluations, including accuracy (\emph{ACC}), precision (\emph{PRE}), recall (\emph{REC}), and F1-score (\emph{F1}). To tackle the class imbalance problem, we trained all evaluated methods by using the oversampling technique. To reduce bias, we repeated running $20$ times and reported the average performance. In addition, the configurations of all evaluated methods were shown in Appendix~\ref{sec:expesettings}.

%%%%%%%%%%%%%%%%%%   模型性能比较   %%%%%%%%%%%%%%%%%
\begin{table*}[htbp]
\centering
\resizebox{\linewidth}{!}{
    \begin{tabular}{c|c@{\hspace{3pt}}c@{\hspace{3.2pt}}c@{\hspace{3.2pt}}c@{\hspace{3.2pt}}|c@{\hspace{3.2pt}}c@{\hspace{3.2pt}}c@{\hspace{3.2pt}}c@{\hspace{3.2pt}}|c@{\hspace{3.2pt}}c@{\hspace{3.2pt}}c@{\hspace{3.2pt}}c@{\hspace{3.2pt}}}
    \toprule[1.2pt]
    \multirow{2}{*}{Methods} & \multicolumn{4}{c|}{\cellcolor{gray!15}\textbf{CLDInst}}     & \multicolumn{4}{c|}{\cellcolor{gray!15}\textbf{Clickbait17}}     & \multicolumn{4}{c}{\cellcolor{gray!15}\textbf{FakeNewsNet}}   \\
    & ACC $\uparrow$    & PRE $\uparrow$    & REC $\uparrow$    & F1 $\uparrow$     & ACC $\uparrow$    & PRE $\uparrow$    & REC $\uparrow$    & F1 $\uparrow$     & ACC $\uparrow$    & PRE $\uparrow$    & REC $\uparrow$    & F1 $\uparrow$     \\ \midrule[0.75pt]
    
    % \textbf{SVM-TS}
    % & 75.38\tiny{$\pm$ 0.18} & \underline{79.45}\tiny{$\pm$ 0.07} 
    % & 72.67\tiny{$\pm$ 0.13} & 75.91\tiny{$\pm$ 0.33} &
    % 69.35\tiny{$\pm$ 0.31} & 72.45\tiny{$\pm$ 0.27} & 67.62\tiny{$\pm$ 0.23} & 69.95\tiny{$\pm$ 0.18} & 71.53\tiny{$\pm$ 0.08} & 73.12\tiny{$\pm$ 0.29} & 69.33\tiny{$\pm$ 0.12} & 71.17\tiny{$\pm$ 0.37}  \\
    % \textbf{BiLSTM} 
    % & 68.21\tiny{$\pm$ 0.23} & 71.23\tiny{$\pm$ 0.16} & 65.45\tiny{$\pm$ 0.19} & 68.23\tiny{$\pm$ 0.28} & 73.78\tiny{$\pm$ 0.20} & 76.34\tiny{$\pm$ 0.05} & 71.21\tiny{$\pm$ 0.09} & 73.70\tiny{$\pm$ 0.24} & 74.26\tiny{$\pm$ 0.34} & 77.45\tiny{$\pm$ 0.11} & 72.56\tiny{$\pm$ 0.09} & 74.93\tiny{$\pm$ 0.15}  \\
    
    % \textbf{dEFEND}       
    % & 73.58\tiny{$\pm$ 0.06} & 77.23\tiny{$\pm$ 0.12} & 72.67\tiny{$\pm$ 0.29} & 74.88\tiny{$\pm$ 0.37} & 74.51\tiny{$\pm$ 0.33} & 78.56\tiny{$\pm$ 0.08} & 72.34\tiny{$\pm$ 0.27} & 75.32\tiny{$\pm$ 0.10} & 75.86\tiny{$\pm$ 0.21} & 79.23\tiny{$\pm$ 0.17} & 73.45\tiny{$\pm$ 0.36} & 76.23\tiny{$\pm$ 0.28}  \\

    \textbf{dEFEND}       
    & 76.58\tiny{$\pm$ 0.06} & 74.03\tiny{$\pm$ 0.12} & 76.46\tiny{$\pm$ 0.29} & 75.23\tiny{$\pm$ 0.37} & 82.15\tiny{$\pm$ 0.33} & 76.74\tiny{$\pm$ 0.08} & 79.22\tiny{$\pm$ 0.27} & 77.96\tiny{$\pm$ 0.10} & 81.06\tiny{$\pm$ 0.21} & 74.21\tiny{$\pm$ 0.17} & 78.74\tiny{$\pm$ 0.36} & 76.41\tiny{$\pm$ 0.28}  \\
    
    \textbf{HPFN}
    & 73.15\tiny{$\pm$ 0.14} & 72.12\tiny{$\pm$ 0.31} & 70.23\tiny{$\pm$ 0.27} & 71.16\tiny{$\pm$ 0.15} &
    81.68\tiny{$\pm$ 0.26} & 84.25\tiny{$\pm$ 0.13} & 82.35\tiny{$\pm$ 0.34} & 83.29\tiny{$\pm$ 0.22} & 84.82\tiny{$\pm$ 0.17} & 84.43\tiny{$\pm$ 0.13} & 85.68\tiny{$\pm$ 0.24} & 85.05\tiny{$\pm$ 0.23}  \\

    \textbf{MCAN}
    & 79.21\tiny{$\pm$ 0.18} & 77.45\tiny{$\pm$ 0.07} & 77.80\tiny{$\pm$ 0.19} & 77.62\tiny{$\pm$ 0.28} &
    84.51\tiny{$\pm$ 0.20} & 85.56\tiny{$\pm$ 0.09} & 82.74\tiny{$\pm$ 0.33} & 84.13\tiny{$\pm$ 0.14} & 83.82\tiny{$\pm$ 0.29} & 85.13\tiny{$\pm$ 0.37} & 82.21\tiny{$\pm$ 0.18} & 83.64\tiny{$\pm$ 0.15}  \\

    \textbf{CPDM}
    & 80.04\tiny{$\pm$ 0.23} & 77.89\tiny{$\pm$ 0.42} & 79.31\tiny{$\pm$ 0.11} & 78.59\tiny{$\pm$ 0.39} &
    86.14\tiny{$\pm$ 0.07} & 86.30\tiny{$\pm$ 0.10} & 83.07\tiny{$\pm$ 0.21} & 84.65\tiny{$\pm$ 0.28} & 84.27\tiny{$\pm$ 0.13} & 85.18\tiny{$\pm$ 0.22} & 82.44\tiny{$\pm$ 0.14} & 83.79\tiny{$\pm$ 0.31}  \\
    
    \textbf{CCD}
    & 82.77\tiny{$\pm$ 0.14} & 83.13\tiny{$\pm$ 0.10} & 82.91\tiny{$\pm$ 0.16} & 83.02\tiny{$\pm$ 0.24} &
    88.36\tiny{$\pm$ 0.22} & \underline{87.61}\tiny{$\pm$ 0.15} & \underline{86.46}\tiny{$\pm$ 0.09} & \underline{87.03}\tiny{$\pm$ 0.27} & 87.72\tiny{$\pm$ 0.13} & 85.96\tiny{$\pm$ 0.20} & \underline{86.46}\tiny{$\pm$ 0.15} & 86.21\tiny{$\pm$ 0.14}  \\
    
    \textbf{VLP}       
    & \underline{85.56}\tiny{$\pm$ 0.16} & \underline{84.25}\tiny{$\pm$ 0.07} & \underline{83.87}\tiny{$\pm$ 0.20} & \underline{84.06}\tiny{$\pm$ 0.12} & \underline{88.70}\tiny{$\pm$ 0.26} & 87.34\tiny{$\pm$ 0.37} & 86.02\tiny{$\pm$ 0.41} & 86.67\tiny{$\pm$ 0.24} & \underline{88.02}\tiny{$\pm$ 0.10} & \underline{87.25}\tiny{$\pm$ 0.16} & 86.23\tiny{$\pm$ 0.19} & \underline{86.74}\tiny{$\pm$ 0.30}  \\ \midrule[0.75pt]
    
    \textbf{Ours}          
    & \textbf{88.79}\tiny{$\pm$ 0.11} & \textbf{88.73}\tiny{$\pm$ 0.04} & \textbf{89.16}\tiny{$\pm$ 0.32} & \textbf{88.94}\tiny{$\pm$ 0.14} & \textbf{92.83}\tiny{$\pm$ 0.27} & \textbf{93.45}\tiny{$\pm$ 0.17} & \textbf{93.59}\tiny{$\pm$ 0.15} & \textbf{93.52}\tiny{$\pm$ 0.20} & \textbf{91.56}\tiny{$\pm$ 0.07} & \textbf{92.74}\tiny{$\pm$ 0.38} & \textbf{92.97}\tiny{$\pm$ 0.13} & \textbf{92.85}\tiny{$\pm$ 0.08}  \\ \bottomrule[1pt]
    \end{tabular}
    }
\caption{Comparisons of all methods. The improvements were significant using a statistic t-test with p-value$<$0.005.}
\label{tab:comparison}
\end{table*}  

\subsection{Comparisons against State-of-the-arts}\label{sec:comparesota}
To verify the effectiveness of our method, we compared it against six typical baselines in the field of clickbait detection. including (1) \emph{dEFEND}~\cite{shu2019defend}, a co-attention-based model that predicted based on both post content and user profiles; (2) \emph{HPFN}~\cite{shu2020hierarchical}, which made judgments based on posts' propagation on social network; (3) \emph{MCAN}~\cite{wu2021multimodal}, a multi-modal model that captured both textual and visual features by stacked co-attention layers; (4) \emph{CPDM}~\cite{DBLP:journals/corr/abs-2112-08611}, using ensemble classifier to perceive inconsistencies among content, headline and thumbnail; (5) \emph{CCD}~\cite{chen2023causal}, a model based on causal intervention and counterfactual reasoning; (6) \emph{VLP}~\cite{wang2023all}, a multimodal pre-trained detector.

\begin{figure*}[t]
\centering
\includegraphics[width=\linewidth]{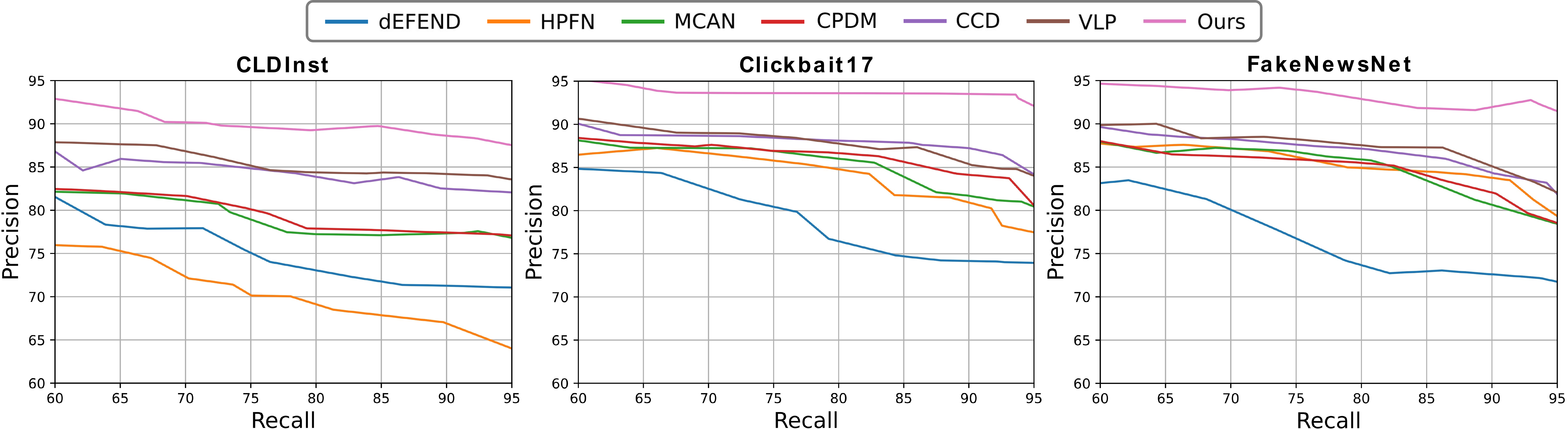}
\caption{PR curves of all models on three datasets.}
\label{fig:pr_curve}
\end{figure*}

As displayed in Table~\ref{tab:comparison}, our method achieved the best performance. The outperformance was over the best baselines (e.g., \emph{VLP}) on \emph{CLDInst}, \emph{Clickbait17}, and \emph{FakeNewsNet} by {3.78\%}, {4.66\%}, and {4.02\%} in terms of the accuracy metric, respectively. Methods with multimodal features (i.e., \emph{MCAN}, \emph{CPDM}, \emph{CCD} and \emph{VLP}) showed better performance, since these features provided useful discriminant clues. In addition, our method performed better than the causal baseline \emph{CCD}. \emph{CCD} only tackled the misalignment among the textual and visual features but neglected the spurious bias and disguised content that are widespread in bait posts. Besides, we observed on the larger datasets e.g.,\emph{Clickbait17} and \emph{FakeNewsNet}, our outperformance was bigger. The reason may be that spurious bias in these datasets was more extensive, and our debiased gain was greater. To evaluate it, we further selected 500 test samples randomly from each dataset and annotated each post manually. We found that approximately $23\%$, $26\%$, $27\%$ of the posts were disguised type, respectively. The datasets \emph{Clickbait17} and \emph{FakeNewsNet} were more complicated, having a larger percentage of disguised samples. Moreover, we provided the precision-recall curves on three datasets in Fig.(\ref{fig:pr_curve}). Our model achieved the best precision across all recall levels. That reflected the effectiveness of our approach in eliminating spurious bias. Ours can obtain a high recall rate and well identify the bait posts dressed up as valid ones.

%%%%%%%%%%%%%%%%% 这是PR曲线 %%%%%%%%%%%%%%%%%

\subsection{Ablation Studies}\label{sec:ablation}
To gain insight into the relative contributions of each component in our approach, we performed ablation studies on four aspects, including (1) \emph{w/o MF} that discarded the multimodal representation module and relied solely on text encoders; (2) \emph{w/o EICF} that dropped the invariant causal factor by removing the \emph{IRM} regularization in Eq.(\ref{formula:opobj}); (3) \emph{w/o ENF} which threw away the scenario learning module, randomly assigned scenarios to samples; (4) \emph{w/o ESCF} that removed scenario-specific causal factor by deleting the module in Eq.(\ref{formula:contrastive_loss}).

As shown in Table~\ref{tab:ablation}, the ablation on all evaluated components led to a significant performance drop. This reflected that all four modules were indispensable. Among them, discarding the  \emph{EICF} module caused the most substantial decrease, i.e., more than 6.33\% and 6.77\% drops in terms of the accuracy and F1 metrics, respectively. That indicated the usefulness of eliminating spurious bias, which can improve discriminant and generalization ability. The multimodal module was also impactful, with its removal leading to a reduction of around 4.61\% $\sim$ 5.63\% in terms of accuracy. This demonstrated the benefits of inter-modal signals to overcome unimodal one-sidedness. In addition, the \emph{ENF} and \emph{ESCF} modules focus on scenario-specific and non-causal factors, which also led to noticeable performance drops when they were ablated. The results validated the rationality of our design. Ablation studies on other metrics were shown in Appendix~\ref{sec:ablation_other}.
%%%%%%%%%%%%%%%%%%% Ablation Study Table %%%%%%%%%%%%%%%%%%%
\begin{table}[h]
\centering
\resizebox{\linewidth}{!}{
\begin{tabular}{c@{\hspace{5pt}}|c@{\hspace{4.8pt}}c@{\hspace{4.8pt}}|c@{\hspace{4.8pt}}c@{\hspace{4.8pt}}|c@{\hspace{4.8pt}}c@{\hspace{4.8pt}}}
\toprule[1.2pt]
\multirow{2}{*}{Datasets}  & \multicolumn{2}{c|}{\cellcolor{gray!15}\textbf{CLDInst}} & \multicolumn{2}{c|}{\cellcolor{gray!15}\textbf{Clickbait17}} & \multicolumn{2}{c}{\cellcolor{gray!15}\textbf{FakeNewsNet}} \\

        & ACC $\downarrow$   & F1 $\downarrow$ & ACC $\downarrow$   & F1 $\downarrow$ & ACC $\downarrow$   & F1  $\downarrow$  \\ \midrule[0.8pt]

w/o MF   
& -4.09\tiny{$\pm$ 0.10}  & -4.72\tiny{$\pm$ 0.08} & -5.23\tiny{$\pm$ 0.13}   & -5.96\tiny{$\pm$ 0.06} & -4.81\tiny{$\pm$ 0.09}   & -5.60\tiny{$\pm$ 0.07}             \\

w/o EICF 
& \textbf{-5.62}\tiny{$\pm$ 0.14}   & \textbf{-6.03}\tiny{$\pm$ 0.06} & \textbf{-7.22}\tiny{$\pm$ 0.09}    & \textbf{-8.06}\tiny{$\pm$ 0.12} & \textbf{-6.62}\tiny{$\pm$ 0.18}  & \textbf{-7.07}\tiny{$\pm$ 0.09}         \\

w/o ESCF 
& -2.88\tiny{$\pm$ 0.05}   & -3.70\tiny{$\pm$ 0.08} & -3.86\tiny{$\pm$ 0.16}  & -4.27\tiny{$\pm$ 0.11}  & -3.64\tiny{$\pm$ 0.03}  & -3.87\tiny{$\pm$ 0.14}   \\

w/o ENF  
& -3.33\tiny{$\pm$ 0.06}   & -3.92\tiny{$\pm$ 0.10} & -4.11\tiny{$\pm$ 0.18}  & -4.73\tiny{$\pm$ 0.09}  & -3.77\tiny{$\pm$ 0.15}  & -4.31\tiny{$\pm$ 0.05}   \\ \bottomrule[1.0pt]
\end{tabular}}
\caption{Ablation study with t-test, p-value$<$0.005.}
\label{tab:ablation}
\end{table}

% Moreover, we evaluated the usefulness of the mask mechanism and scenario in our model to learn debiased representation. The results were shown in Appendix~\ref{sec:maskstudy} and Appendix~\ref{sec:scenario}, respectively.

\subsection{Study of the Mask Mechanism}
\begin{figure}[h]
\centering
\includegraphics[width=\linewidth]{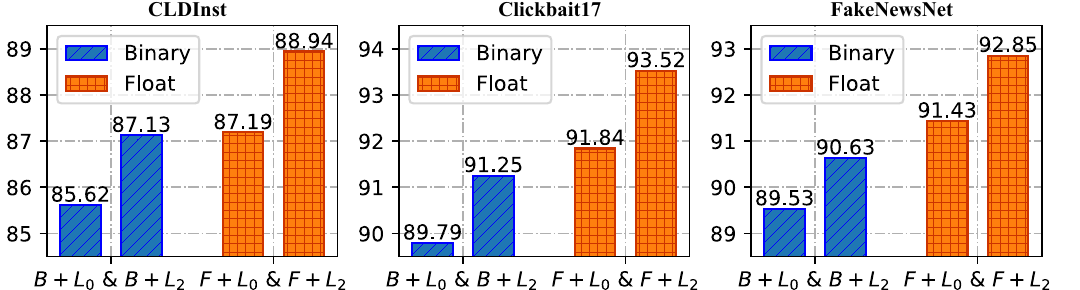}
\caption{Evaluation the impact of $\mathbf{m}$ settings on F1.}
\label{fig:mask study}
\end{figure}

Relying on the mask vector $\mathbf{m}$, our model can extract a causal invariant factor applicable to most scenarios. We observed that the setting of $\mathbf{m}$ can be binary ($B$) or float ($F$), and it can be integrated into the objective Eq.(\ref{formula:opobj}) by the $L_2$ norm or a $L_0$ regularizer. To better understand how the mask $\mathbf{m}$ impacted the performance, we tested four $\mathbf{m}$ configurations: ($B+L_0$), ($B+L_2$), ($F+L_0$) and ($F+L_2$). As shown in Fig.(\ref{fig:mask study}), we found that the float masks (F) consistently outperformed binary masks (B). This verified that keeping the continuous mask values provided more representational power than binary ones. In addition, $L_2$ regularization worked better than $L_0$. Among all datasets, $L_2$ norm outperformed $L_0$ norm by around 1.23\% $\sim$ 2.01\% on the F1 score. The sparse $L_0$ regularization might suppress informative dimensions, while $L_2$ allowed more flexibility. More evaluations on the mask mechanism were shown in Appendix~\ref{sec:maskstudy}.
%%%%%%%%%%%%%%%%%%% Study of Mask %%%%%%%%%%%%%%%%%%%

\subsection{Study of the Scenario}

\begin{figure}[h]
\centering
\includegraphics[width=\linewidth]{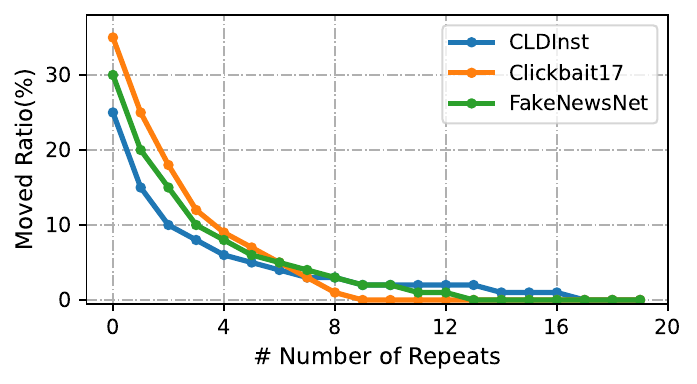}
\caption{Evaluation on scenario separability based on the ratio of moved samples.}
\label{fig:envimpact-1}
\end{figure}

Scenario is valuable to learn invariant representations. These stable features can effectively improve the robustness of the model and avoid being deceived by disguised content. To study the characteristics of scenarios, we first checked their separability. It was estimated from the data based on alternating optimization. That is, the samples were grouped to form new scenarios; in turn, after the scenarios were updated, the samples were moved to partition again. If most samples no longer changed, the scenarios can be viewed to be divisible. Thus, we calculated the ratio of moved samples to infer the stability of scenarios. If a sample cannot fit the current scenario and needed to be reassigned to a new one, we counted it as a moved case. As shown in Fig.(\ref{fig:envimpact-1}), the curves of the moved rate on three datasets converged after being repeated around $10$ rounds. That indicated the scenarios can be separated and help to capture the spurious bias well. Moreover, the curve implied the appropriate loop count for the model optimization. Moreover, we evaluated the scenario size setting in Appendix~\ref{sec:scenario}.
%%%%%%%%%%%%%%%%%%%% study of the scenarios figures %%%%%%%%%%%%%%%%%%%%%

\subsection{Case Studies}\label{sec:casestudy}

\begin{figure}[t]
\centering
\includegraphics[width=\linewidth]{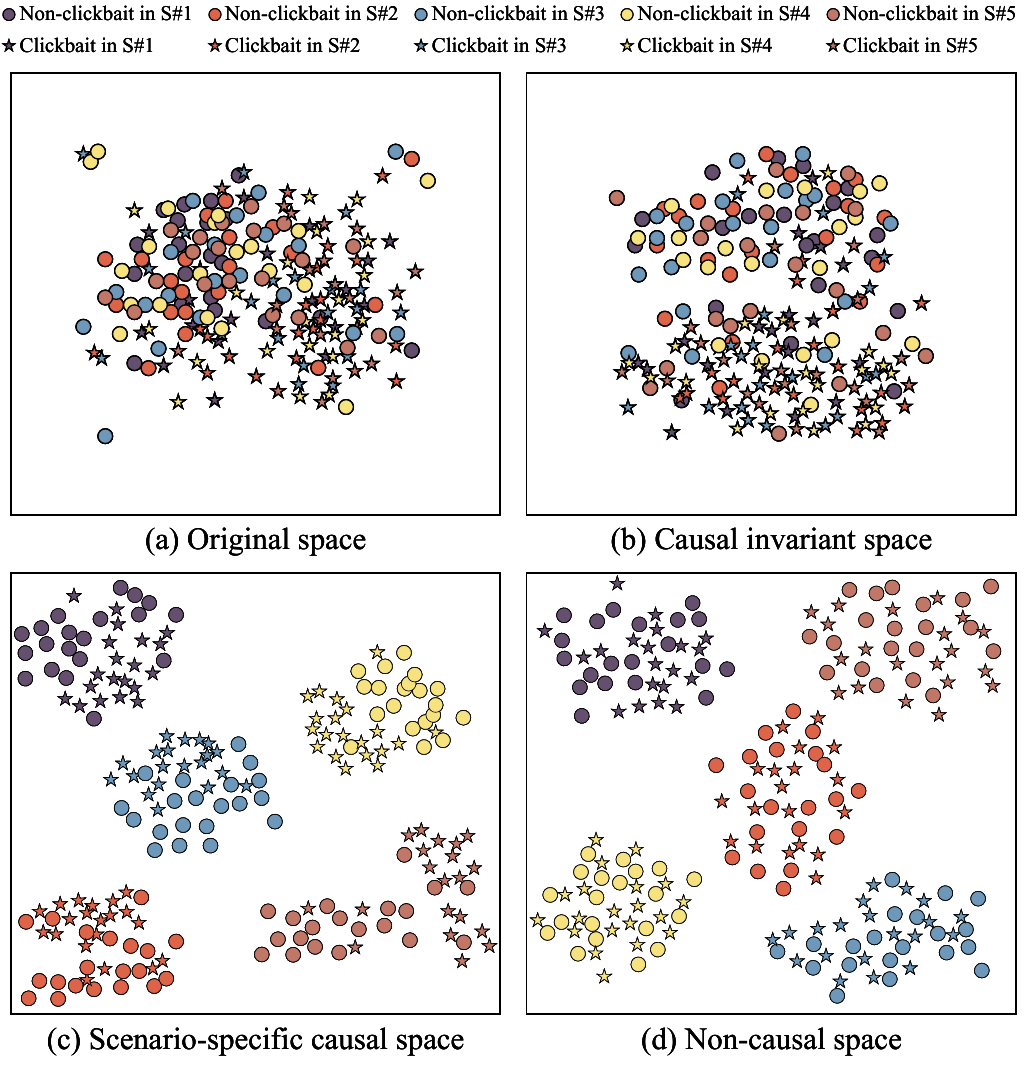}
\caption{Case study with t-SNE visualization on the distribution of samples in several typical scenarios.}
\label{fig:case_study}
\end{figure}

Furthermore, we conducted case study to see the actual effect of our model. Given a dataset like \emph{Clickbait17}, we predicted a score for each test sample. The highest score indicated the scenario it belonged to. After classifying all samples, we selected 5 scenario sets with the largest size of samples. We randomly chose 20 bait and 20 non-bait samples from each scenario, and visualized their features by the dimension reduction tool \emph{t-SNE}. As shown in Fig.(\ref{fig:case_study}), subgraph (a) showed the original features cannot differentiate scenarios; (b) presented the causal invariant features learned by our model. These features would be helpful for prediction, but they could not distinguish scenarios. The samples formed loosely separated clusters but remained disorganized within each cluster; (c) visualized the scenario-specific causal features. Here, scenarios were separated into distinct clusters well. The features can be roughly divided into two groups, but the inter-class discrimination was insufficient. This demonstrated the need to combine invariant and scenario-specific causal factors for robust prediction; (d) depicted a noise factor. Since the noises in each scenario are unique, they may separate the scenarios but it is weak to identify true/false samples. Overall, these results demonstrated how our model elicited key causal factors and discarded spurious correlations, which enabled reliable prediction.

\section{Related Work}
\label{sec:related-work}

Clickbait detection is a hot research topic that aims to predict misleading and sensational social posts. The predictive sources can be divided into two categories. The first one is based on social activity. Malicious creators keep posting clickbait and ceaselessly evolve new subspecies to avoid detection. Some researchers propose to examine this bait behavior based on the creators' account profiles~\cite{yang2012automatic} and activity metadata~\cite{shu2019role}. Nevertheless, some valid creators would produce both high-quality and poor posts, which easily led to false positive predictions. On the other hand, clickbait posts often need to be widely spread on social networks to enhance their influence. These posts should have remarkable propagation characteristics, such as a unique spread path, a large number of shares but a short reading time due to dissatisfaction, etc. The propagation-based methods had been proposed~\cite{ma2017detect}, but they cannot tackle new posts due to lack of user feedback~\cite{ma2018rumor}. Thus, such methods often perform high accuracy, but low coverage. Also, they have an alert delay, which would bring a lot of losses and could not support online applications. 

Another direction is based on the post's content. The bait posts often have certain language characteristics, such as delusive words, fallacious phases, hot topics, raged emotions~\cite{guo2019exploiting}, linguistic stylometry~\cite{potthast2017stylometric}, etc. To detect them, early works design rules based on various features, such as the semantic~\cite{DBLP:conf/asunam/RonyHY17}, linguistic~\cite{blom2015click}, or multimodal ones~\cite{DBLP:conf/icmi/ChenCR15}. However, these rules are hand-crafted and non-extensible~\cite{DBLP:conf/www/YuQS020}. To tackle this issue, current research turns to data-driven neural networks. Various network architectures have been proposed~\cite{krishneth2023web}, such as \emph{RNN}~\cite{DBLP:conf/ecir/Anand0P17}, \emph{CNN}~\cite{CNN-Agrawal}. More advanced techniques like feature attention~\cite{indurthi-etal-2020-predicting} and graph attention mechanisms~\cite{DBLP:journals/kbs/LiuYWZZW22} have been developed. To learn representations from rich unlabeled data~\cite{DBLP:journals/tkde/YuSQY23}, some researchers explore the pre-trained language models (\emph{PLMs}), like \emph{BERT}~\cite{Devlin2018} and \emph{RoBERTa}~\cite{DBLP:journals/corr/abs-1907-11692}. They are fine-tuned to fit the classified tasks~\cite{indurthi-etal-2020-predicting}, so as to use their embedded semantic knowledge~\cite{DBLP:conf/acl/Yu0ZSLZ023} to facilitate detection~\cite{ijcai2022p621}. Besides, some works point out that the clickbait was not only in the unimodal like the text~\cite{ruchansky2017csi} but also in multimodal~\cite{shu2019defend}, such as giving an attractive image with an unrelated article. They propose to incorporate more features in multiple modalities~\cite{wang2018eann}. For model training, there are other studies on domain adaptation~\cite{lopez2018hybridizing} and data augmentation~\cite{yang2019fake} to solve the data shortage problem by knowledge transfer.

Differently, we found that traditional representations have spurious correlations. Malicious creators would exploit this bug to yield a large number of new subspecies by rewriting the posts' content, such as adding some valid but unrelated terms or images, and concealing the bait content in a normal post, etc. In this mixed content environment, the existing model easily misjudges the bait posts due to spurious bias. We thus propose a debiased method to tackle this problem by causal inference. 
 
Causality-inspired methods are a hot research topic in many tasks~\cite{nguyen2022front}. To grasp the causality in the tasks, some works proposed the causal loss function~\cite{bagi2023generative}, while others designed a task-related causal structure~\cite{liu2022towards} to guide decision-making~\cite{lv2022causality}. They pointed out the value of a good representation for the model's performance~\cite{bronakowski2023automatic}. Some studies propose to capture the salient representations by feature engineering~\cite{yu2018aesthetic} or noise filtering~\cite{wang2021clicks}, but they ignore spurious correlations. In contrast, we propose to elicit key factors to eliminate this bias.

\section{Conclusion}

This paper studied the task of detecting clickbait in multimodal social media posts. Existing methods learned shallow features that introduced spurious correlations, rather than capturing inherent factors that caused clickbait. Malicious creators used this spurious bias to form new bait subspecies by rewriting the posts with tricks, such as disguise with valid content, leading to misjudgment and poor robustness. To tackle this problem, we proposed a new debiased framework by causal representation inference. In detail, we first extracted multimodal features, including textual, visual, linguistic, cross-modal, and creator profile features. By causal inference with structural constraints, we then disentangle them into three latent factors, including the invariant factor that indicated inherent bait intentions, a causal factor for a certain scenario, and noise factor. Based on invariant and causal factors, we can build a robust model. Moreover, we propose a data augmentation technique to reduce training costs. Experimental results on three popular datasets shown the effectiveness of our model.

\section*{Acknowledgments}
This work is supported by the National Natural Science Foundation of China (62276279, 62372483, 62472455, 62102463, U2001211, U22B2060), Guangdong Basic and Applied Basic Research Foundation (2024B1515020032), Research Foundation of Science and Technology Plan Project of  Guangzhou City (2023B01J0001, 2024B01W0004), and Tencent WeChat Rhino-Bird Focused Research Program (WXG-FR-2023-06).

\section*{Limitations}

Detecting clickbait posts on social media is a challenging task. In real application, this task comes up against issues such as biases, multimodal content, evolving bait subspecies, few labeled resources, etc. Moreover, malicious creators would constantly disguise the bait posts with tricks such as replacing confusing terms or images pictures to escape detection. This replaced content is full of spurious correlations which would make the model misjudgment. That would seriously affect the model's robustness. We thus propose a debiased model, which can find the posts with inconsistency on the thumbnail-article and headline-article pairs, such as this thumbnail and headline are catchy and sensational, but irrelevant to the article. Although the current model is effective, there is still room for improvement. For example, it does not verify whether the events described in the post are true or fake. Also, it does not cover the detection of video bait posts. Fake content detection is another task and remains an open challenge. A possible solution is to build a trusted knowledge base. We will investigate more data modalities in future work.

\section*{Ethics Statement}

The technology proposed in this paper can be used to filter the bait posts. That can improve the user experience and purify the online environment. Unlike traditional methods based on shallow features, our model is more robust by eliminating spurious bias. Excluding the misusage scenarios, there are few or even no ethical issues with this technology. However, it is essentially a classification method, and the classified results may be misused. For example, it may be abused by malicious persons to filter out the posts created from certain commercial competitors unfairly. This problem can be addressed by analyzing the source and publisher of the posts.

\bibliography{custom}

\appendix

\section{Multimodal Feature Extraction} \label{sec:linguistic}

\textbf{Visual Features}: \emph{Swin-T}~\cite{liu2021swin}  is a transformer-based model that is good at capturing hierarchical features in images based on shifted window self-attention. \\
\textbf{Cross-modal Features}: We employ a \emph{CT} Transformer~\cite{lu2019vilbert} to compute the cross-modal features between the text and image.  Given the textual encoding $\mathbf{T}^b$ and visual one $\mathbf{V}^s$ with their data type tags, \emph{CT} can well encode their matching features by a multi-headed attention network. It outputs a visual-aware text feature $\mathbf{F}^{vt}$ and a text-aware visual feature $\mathbf{F}^{tv}$, respectively by $\mathbf{F}^{vt} = CT((\mathbf{T}^b \mathbf{W}^t), (\mathbf{V}^s \mathbf{W}^v))$, $\mathbf{F}^{tv} = CT((\mathbf{V}^s \mathbf{W}^v), (\mathbf{T}^b \mathbf{W}^t))$, where $\mathbf{W}$ are weight matrices. These encodings can better reflect bait behavior from multiple views.\\
\textbf{Linguistic Features}: To capture the semantics and bait patterns of the post, we extract six kinds of features based on post content, i.e., thumbnail, headline, and article body:\\
(1) \textit{ArticleBody-thumbnail disparity}: Creators may exploit tricks, such as using attractive thumbnails that are irrelevant to the article body, to mislead readers into clicking.
To calculate this cross-modal inconsistency, we encode them into a uniform mathematical space by a pre-trained \emph{CLIP} model~\cite{radford2021learning}, as $[b^c, t^c] = CLIP(b, t)$, where $b$ denotes the body text or figure in the article, $t$ refers to the thumbnail picture, $b^c$ and $t^c$ are their encoded vectors, respectively, where $d_c$ is the length. Previous works~\cite{li2022languagedriven} show that \emph{CLIP} has a good ability to characterize the mutual relations of various modalities.\\
(2) \textit{ArticleBody-headline disparity}: To attract readers, creators often use lure headlines to form a curiosity gap, regardless of their relevance to the articles. To compute their relevance, we first embed each textual input via \emph{BERT} and encode its context by a bidirectional \textit{GRU} with an attention layer. We then feed these two inputs into a \emph{Siamese} net~\cite{neculoiu2016learning} that is good at learning similarity, as $sim_{bh} = Siamese(b, h)$, where $b$ denotes the body text, $h$ refers to the headline.\\
(3) \textit{Thumbnail-headline disparity}: When the headline doesn't match the thumbnail, it will create a curiosity gap and cause misleading clicks. To understand the thumbnail, we generate its caption based on a neural model~\cite{xu2015show} trained on \emph{MS COCO}~\cite{lin2014microsoft} dataset. We then compare the generated caption and headline by cosine matching of their \emph{BERT} embeddings.\\
(4) \textit{Sentiment of headline}: The headline often expresses strong feelings to form an emotional curiosity gap. To analyze such feelings, we use a sentiment classifier~\cite{hossain2021text} which outputs both polarity (positive/negative/neutral) and intensity (strength). We normalize their values to the range $[0,1]$ and obtain a feature vector.\\
(5) \textit{Lexical analysis of headline}: To grab the reader's attention, the headline usually uses provoking clues, such as the overuse of numbers, question marks, exclamation marks, and capital letters to emphasize the shock, or using emojis to indicate the funny emotion. We thus design a count-based vector to quantify the features like capitalized letters, question marks, punctuation marks such as `$\sim$,!', emojis, assertive verbs,  factive verbs, hedges, implicative verbs, etc.\\
(6) \textit{Baitness of headline}: Another striking peculiarity is the lure verbalism, for example, using abbreviations like ``\emph{OMG}'' i.e., \textit{oh my god} to express surprise, ``\emph{LOL}'' i.e., \textit{laughing out loud} to describe humor, ``\emph{ROFL}'' i.e., \textit{rolling on the floor laughing}. The use of celebrity names or pornographic words~\cite{alcantara2020offensive} like ``\emph{nudes}.'' We thus extract features by counting the number of celebrities, slang words such as ``\emph{OMG, WTF}'', porn words like ``\emph{sexy}'', captivating phrases like ``\emph{you wouldn't believe},'' ``\emph{shocking}.''

\section{Experiment Settings} \label{sec:expesettings} 
For evaluation, we reimplemented each baseline with default settings. For fair comparisons, we conducted ten runs and showed the average results.

\textbf{Ours:} Our model was trained on four 24 GB \textit{Nvidia RTX 3090 GPUs}. Based on \emph{HuggingFace PyTorch} API~\cite{DBLP:conf/emnlp/WolfDSCDMCRLFDS20}, we encoded the images based on the {Swin}$_{Tiny}$ model pre-trained on \emph{ImageNet} with $4$ layers and a hidden dimension of $96$. The transformer had a window size of $7\times7$ and $6$ attention heads. For text encoding, we leveraged the BERT$_{base}$ model with $6$ transformer layers and a hidden size of $768$. The classifier was a $3$-layer \emph{MLP} with \emph{ReLU} activations and a hidden size of $512$. The \emph{Adam} optimizer was used with a learning rate of $2e-5$. We trained for $10$ epochs, with the batch size of $64$. For contrastive learning, we used a projection head with $256$ units and a temperature of $0.1$. The mask generator was a $2$-layer \emph{MLP}. The learning rate for $\mathbf{m}$ was obtained by greedy search on $[0.01, 0.001, 0.0001]$. The trade-off $\alpha$ and $\beta$ were set as $2$ and $0.1$ for \emph{CLDInst}, $1$ and $0.1$ for \emph{Clickbait17} and $1$ and $0.01$ for \emph{FakeNewsNet}, respectively. The number of scenarios $|S|$ was set as $10$ for \emph{CLDInst}, $15$ for \emph{Clickbait17} and $15$ for \emph{FakeNewsNet}, respectively. The iteration $T$ was set as $20$ initially.

% The trade-off $\alpha$ and $\beta$ were tuned in $\{2, 1, 0.1, 0.01, 0\}$. The number of scenarios $|S|$ was tuned in $[1, 5, 10, 15, 20]$. The iteration $T$ was initially set as $20$.

\textbf{dEFEND:} We used pre-trained \emph{GloVe} embeddings of dimension 100 to represent words. We incorporated both bidirectional \emph{GRU} layers for sequential data processing and custom attention mechanisms to focus on relevant parts of the text. The max sentence length and sentence count were set to 120 and 50, respectively. We utilized the \emph{RMSprop} optimizer with a learning rate of 0.001. We trained for $10$ epochs, with the batch size of $20$. 

\textbf{HPFN:} We extracted the features from the hierarchical propagation network, and then employed a series of ensemble classifiers to predict the results, \emph{Gaussian Naive Bayes}, \emph{Logistic Regression} with `\textit{lbfgs}' solver, \emph{Decision Tree}, \emph{SVM} with a linear kernel. Also, we used \emph{Random Forest} with 50 estimators for detailed training and 100 estimators for broader model performance analysis. Data normalization was achieved through \emph{StandardScaler}. We used \emph{Extra Trees Classifier} to build a forest with 100 estimators and computed the feature weighting.

\textbf{MCAN:} Textual features were extracted using the \emph{BERT} model, and visual features from images were obtained through the \emph{VGG-19} network. The model incorporated a novel fusion approach with multiple co-attention layers to effectively integrate textual and visual features. Specifically, the model was designed with `\emph{s-fc}', `\emph{f-fc}', and `\emph{t-fc}' layers. Each had a hidden size of 256, and a `\emph{p-fc}' layer with a hidden size of 35. The dimensions $d$, $m$, and $\textit{dff}$ were set to 256, 4, and 512, respectively. Training was conducted for 100 epochs with early stopping, utilizing \emph{Adam} optimizers. The \emph{VGG-19} and \emph{BERT} parameters were frozen to prevent overfitting. Parameters were optimized by grid searching, with accuracy as the selection criterion.

\textbf{CPDM:} We embedded the input text by the \emph{BERT} model. The dimension of the last hidden layer was set to 768. For the visual input, we embedded it by the \emph{ResNet50} pre-trained model with the output layer dimension of 2048. We then fused these multimodal features and trained an ensemble classifier, i.e., \emph{Random Forest}. It consisted of six base classifiers, i.e., \emph{K-Nearest Neighbours}, \emph{Gaussian Naive Bayes}, \emph{Multi-Layer Perceptron}, \emph{Support Vector Machines }, \emph{Extreme Gradient Boosting}, and \emph{Logistic regression}. We used \emph{Adam} as the optimizer with a learning rate of $10^{-4}$ and a batch size of 128. The training epoch was set to 120.

\textbf{CCD:} We trained the model for 200 epochs. The initial learning rate for the \emph{Adam} optimizer was tuned in [1e-5, 1e-3]. For the confounder dictionary $\mathbf{D}_u \in \mathbb{R}^{N \times d_u}$ , $N$ is 18 (\emph{Anger, Anxiety, Assent, Causation, Certainty, Differentiation, Discrepancy, Feel, Hear, Insight, Negative emotion, Netspeak, Nonfluencies, Positive emotion, Sadness, See, Swear words, Tentative}), and $d_u$ was set to 4. For the scaled dot-product attention, the scaling factor $d_m$ was set to 256. We followed the original settings to tune the trade-off hyperparameters $\alpha$ and $\beta$ by grid search in \{0, 0.1, 0.25, 0.5, 0.75, 1, 2, 3, 4, 5\}. And we finally set $\alpha = 3$ and $\beta = 0.1$.

\textbf{VLP:} Due to the storage limitation, we used the first half of the \emph{YT-Temporal} 180M. We trained the model using the \emph{AdamW} optimizer with a base learning rate of 1e-4 and weight decay of 1e-2. The learning rate was warmed up for 10\% of the training steps and was decayed linearly to zero for the rest of the training. For pre-training, we trained \emph{All-in-one-S} and \emph{All-in-one-B} for 200K steps with a batch size of 8 per GPU. For \emph{All-in-one-Ti}, we pre-trained for 100K steps with a batch size of 16 per GPU. We adopted mixed precision technique to speed up the training process. As the domain gap between pre-train dataset and downstream visual dataset is large, we used batch size of 512 and trained with 100 epochs.

%%%%%%%%%%%%%%%%%%% Study of Mask %%%%%%%%%%%%%%%%%%%
\begin{figure}[t]
\centering
\includegraphics[width=\linewidth]{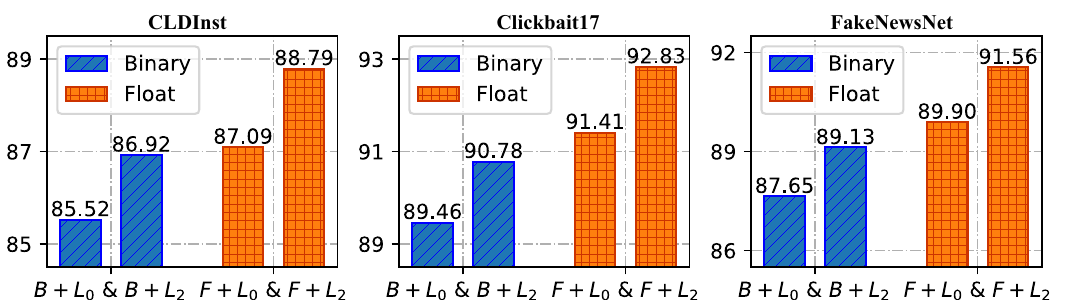}
\caption{Evaluation the impact of $\mathbf{m}$ settings on ACC.}
\label{fig:mask study acc}
\end{figure}

%%%%%%%%%%%%%%%%%%% Study of Mask %%%%%%%%%%%%%%%%%%%
\begin{figure}[t]
\centering
\includegraphics[width=\linewidth]{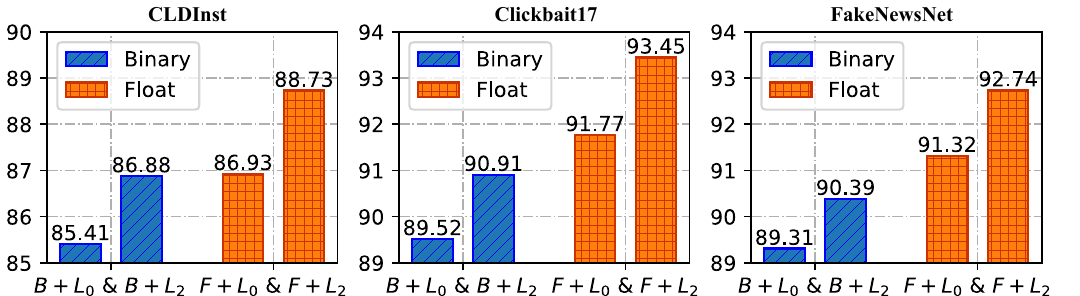}
\caption{Evaluation the impact of $\mathbf{m}$ settings on PRE.}
\label{fig:mask study pre}
\end{figure}

%%%%%%%%%%%%%%%%%%% Study of Mask %%%%%%%%%%%%%%%%%%%
\begin{figure}[t]
\centering
\includegraphics[width=\linewidth]{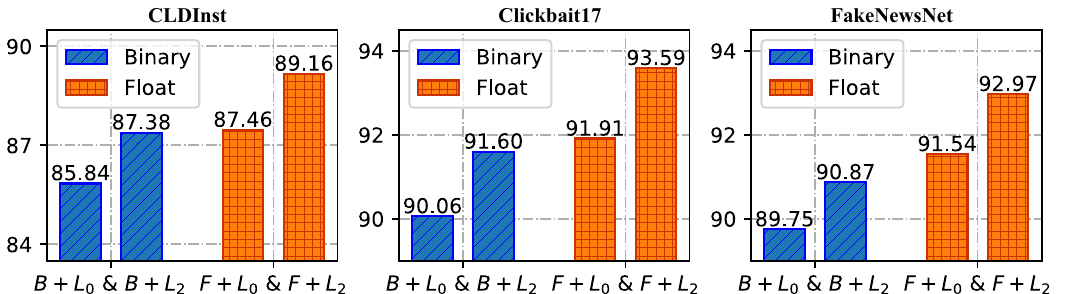}
\caption{Evaluation the impact of $\mathbf{m}$ settings on REC.}
\label{fig:mask study rec}
\end{figure}

%%%%%%%%%%%%%%%%%%% Study of the Mask distribution %%%%%%%%%%%%%%%%%%%
\begin{figure*}[h]
\centering
\includegraphics[width=\linewidth]{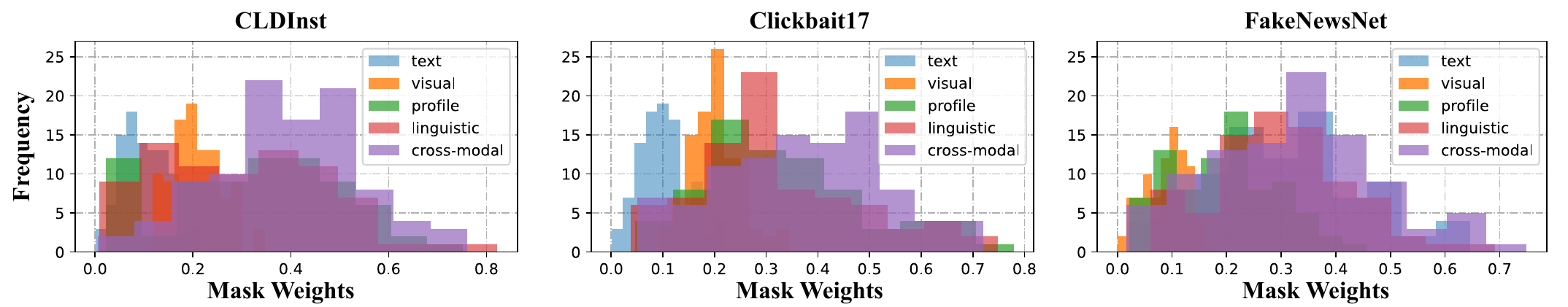}
\caption{Visualization of the mask $\mathbf{m}$ on three datasets. The importance distribution of each kind of feature.}
\label{fig:mask weights}
\end{figure*}

\section{Training Procedure}
\label{sec:training_process}
Our overall training process is summarized in Algorithm~\ref{code:training_process}.

\begin{algorithm}[h]
  \KwData{Training data $\mathcal{D}^{tr}$}
  \KwResult{Optimal detector $\hat{\mathcal{F}}$}
  Random assign scenario for each $x_{i} \in \mathcal{D}^{tr}$\;
  // $T$ rounds of alternating optimization\;
  \For{$t \leftarrow 1$ \KwTo $T$} {
      \While{not converged} {
          Optimize $\Phi_{s}$ via Eq.(\ref{formula:scenario_learning}) for scenario division\;
          Compute $s(i)$ via Eq.(\ref{formula:data_division}) for samples reallocation\;
      }
      Learn \emph{IRM} loss $\mathbf{m}$ via Eq.(\ref{formula:opobj}) to elicit an invariant factor\;
      Optimize the contrastive loss $\xi$ via Eq.(\ref{formula:contrastive_loss}) to separate scenario-specific causal factor and non-causal one\;
   }
   Optimize Eq.(\ref{formula:classifierobj}) to learn a clickbait detector $\hat{\mathcal{F}}$ based on the elicited invariant and causal factors, avoiding spurious bias
\caption{Our model's training process.}
\label{code:training_process}
\end{algorithm}

\section{Additional Evaluations}\label{sec:addevaluations}

Due to the page limit, we showed additional experiments as follows, including the evaluations of the mask and scenario mechanism, case analysis, and data augmentation technique.

\subsection{Additional Study of the Mask Mechanism}
\label{sec:maskstudy}
The results in terms of accuracy, precision, and recall metrics were presented in Fig.(\ref{fig:mask study acc}), Fig.(\ref{fig:mask study pre}), and Fig.(\ref{fig:mask study rec}), respectively. All the results validated our design of using a soft float mask regularized by $L_2$ norm, which could perform best. The mask enabled automatic differentiation of feature relevance for clickbait detection.
%To better understand how the mask $\mathbf{m}$ impacts performance, we conducted experiments with different mask settings. As described above, $\mathbf{m}$ can be binary ($B$) or float ($F$), and regularized by $L_0$ or $L_2$ norm. We tested four $\mathbf{m}$ configurations: ($B+L_0$), ($B+L_2$), ($F+L_0$) and ($F+L_2$). As shown in Figure~\ref{fig:mask study}, we could observe that (1) Float masks (F) consistently outperform binary masks (B). This verified that keeping the continuous mask values provided more representational power than binary ones; (2) $L_2$ regularization worked better than $L_0$. Across all datasets, $L_2$ norm improved over $L_0$ norm by around 0.5-1.5\% on the F1 score. The sparse $L_0$ regularization might suppress informative dimensions, while $L_2$ allowed more flexibility. The results validated our design choice of using a soft float mask regularized by $L_2$ norm, which emerged as the best performer. Furthermore, the mask enabled automatic differentiation of feature relevance for clickbait detection.

To further analyze the mask $\mathbf{m}$, we visualized its per-dimension values in Fig.(\ref{fig:mask weights}). For each dataset, we randomly sampled 100 cases and then analyzed their learned mask weights. This weight captured the importance of each kinds of feature. The value of this weight was between [0,1]. If we plotted the histogram according to the weights of these 100 samples, where the x ordinate is the value of the weight and the y ordinate is its corresponding number of samples. For example, if there was 17 samples had the weight 0.4, then x ordinate was 0.4, the y ordinate was 17. If the weight values of most samples were high, this kind of feature was considered to be more important. From the results, we observed that there were more samples with high weight for the kinds of cross-modal, linguistic, and profile features. This reflected their ability to capture bait behavior more effectively than the single modal features such as text and visual ones. 

%we showed the results in terms of precision, recall the accuracy metrics. 

\subsection{Study of the Scenario Size Setting}
\label{sec:scenario}
\begin{figure*}[t]
\centering
\includegraphics[width=6.1in]{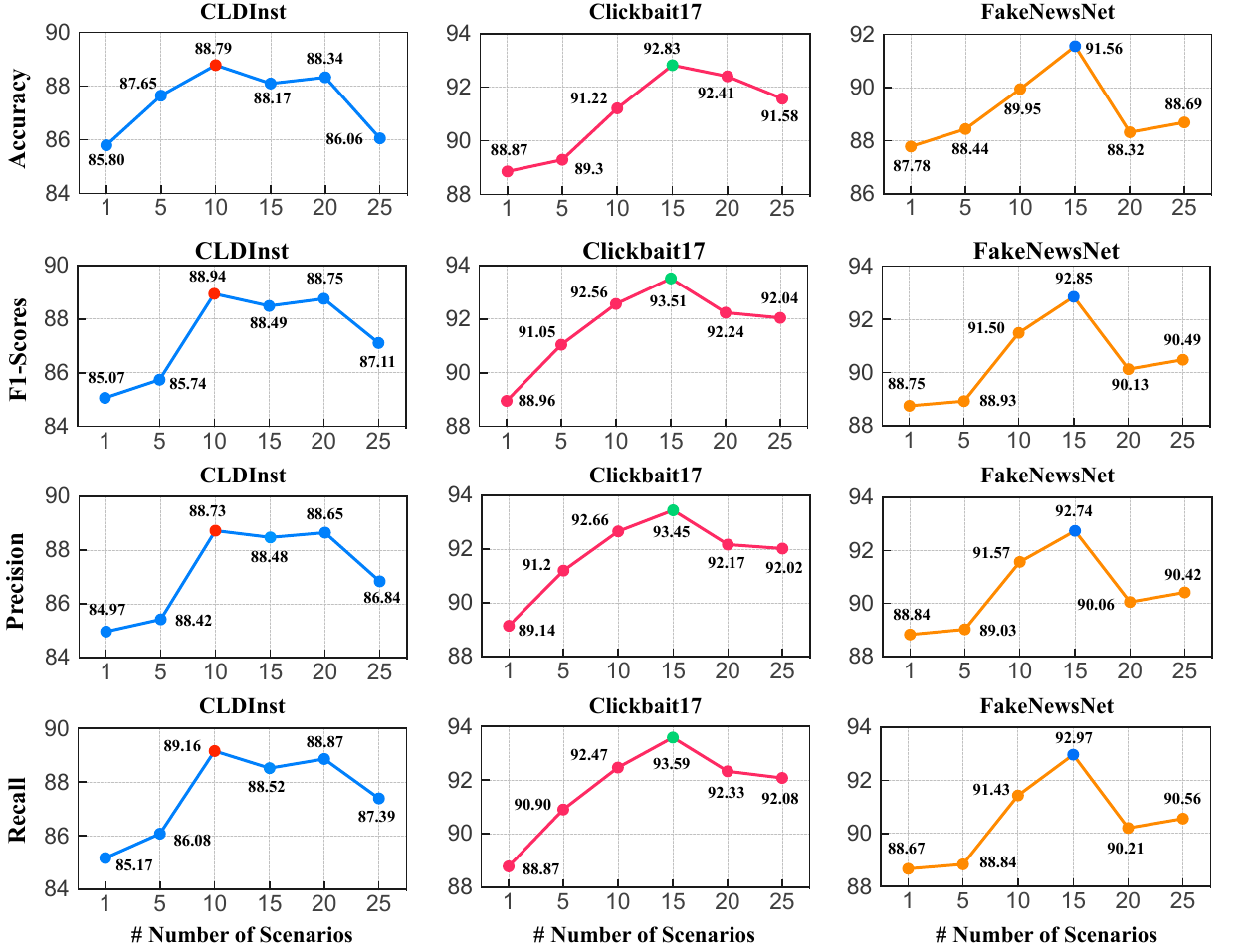}
\caption{Evaluation of scenario size settings in terms of accuracy, F1 score, precision and recall, respectively.}%不同情景大小设置的F1评分比较.Evaluation of the impact of scenario size on the F1 score.%
\label{fig:envimpact-2}
\end{figure*}

We evaluated the impact of scenario size setting on the overall performance. Empirically, the small size was not enough to distinguish complex clickbait patterns, while a large one may split some relevant samples into redundant groups. To explore the best settings, we tuned the scenario size from 1 to 25, with 1 as an interval. Considering the computational resources, we tested the size with no larger than 25. The performance change curves in terms of accuracy, F1 score, precision, and recall were displayed in Fig.(\ref{fig:envimpact-2}). We observed that with the increment of size, the performance had an ascending trend, after reaching the peak, it turned to decline. The optimal sizes for the three datasets were 10, 15, and 15, respectively.

%\begin{figure*}
%\centering
%\includegraphics[width=5.8in]{img/num_scenarios_acc.pdf}
%\caption{Evaluation of the scenario size in terms of ACC.}%不同情景大小设置的ACC评分比较.Evaluation of the impact of scenario size on the F1 score.%
%\label{fig:envimpact-acc}
%\end{figure*}
%
%\begin{figure*}
%\centering
%\includegraphics[width=5.8in]{img/num_scenarios_pre.pdf}
%\caption{Evaluation of the scenario size in terms of PRE.}%不同情景大小设置的PRE评分比较.Evaluation of the impact of scenario size on the PRE score.%
%\label{fig:envimpact-pre}
%\end{figure*}
%
%\begin{figure*}
%\centering
%\includegraphics[width=5.8in]{img/num_scenarios_rec.pdf}
%\caption{Evaluation of the scenario size in terms of REC.}%不同情景大小设置的REC评分比较.Evaluation of the impact of scenario size on the REC score.%
%\label{fig:envimpact-rec}
%\end{figure*}

\subsection{Qualitative Analysis of Test Cases}
\label{sec:predcase}
%%%%%%%%%%%%%%%%%%% case studies table %%%%%%%%%%%%%%%%%%%%%
Moreover, we analyzed the test cases to infer the actual running effect of our model. As exhibited in Table~\ref{tab:case study}, we made the correct prediction for the first post. This article used a cartoon-style thumbnail. Our model could judge the authenticity of this article through the learned textual features and visual consistency, without being misled by the attractiveness of the title or image itself. In the second example, the headline used exaggerated words like ``\emph{Bid Rupert Murdoch farewell by reliving his most controversial tweets}.'' It created a curiosity gap which people expected to reveal the tweets immediately. But the image of the post only contained a side-sitting photo of ``\emph{Rupert Murdoch}'', which had nothing to do with the headline. The body text actually simply reviewed some tweets posted by \emph{Rupert Murdoch} on Twitter, without delivering the implied sensational effect as hinted by the headline. For this clickbait post with inconsistency and irrelevance, other baselines might rely too much on the surface word matching between the headline and body text, without fully understanding their deeper semantics, leading to the wrong judgment. In contrast, our model can work well. These results showed the extracted invariant and scenario-specific factors were effective in distinguishing bait articles. 

\begin{table*}[t]
\footnotesize
\centering
\resizebox{\textwidth}{!}{
\begin{tabular}
{c|c@{\hspace{8.2pt}}|c@{\hspace{8pt}}|c@{\hspace{1.2pt}}|c@{\hspace{1.2pt}}} 
\toprule[1.2pt]
\textbf{Thumbnails} & \centering\textbf{Headlines} & \textbf{Body Texts} & \parbox{1.5cm}{\centering\textbf{True Label}} & \parbox{2cm}{\centering\textbf{Pred Label}} \\ \midrule[0.8pt]

\raisebox{-0.5\height}{\includegraphics[width=1.5cm]{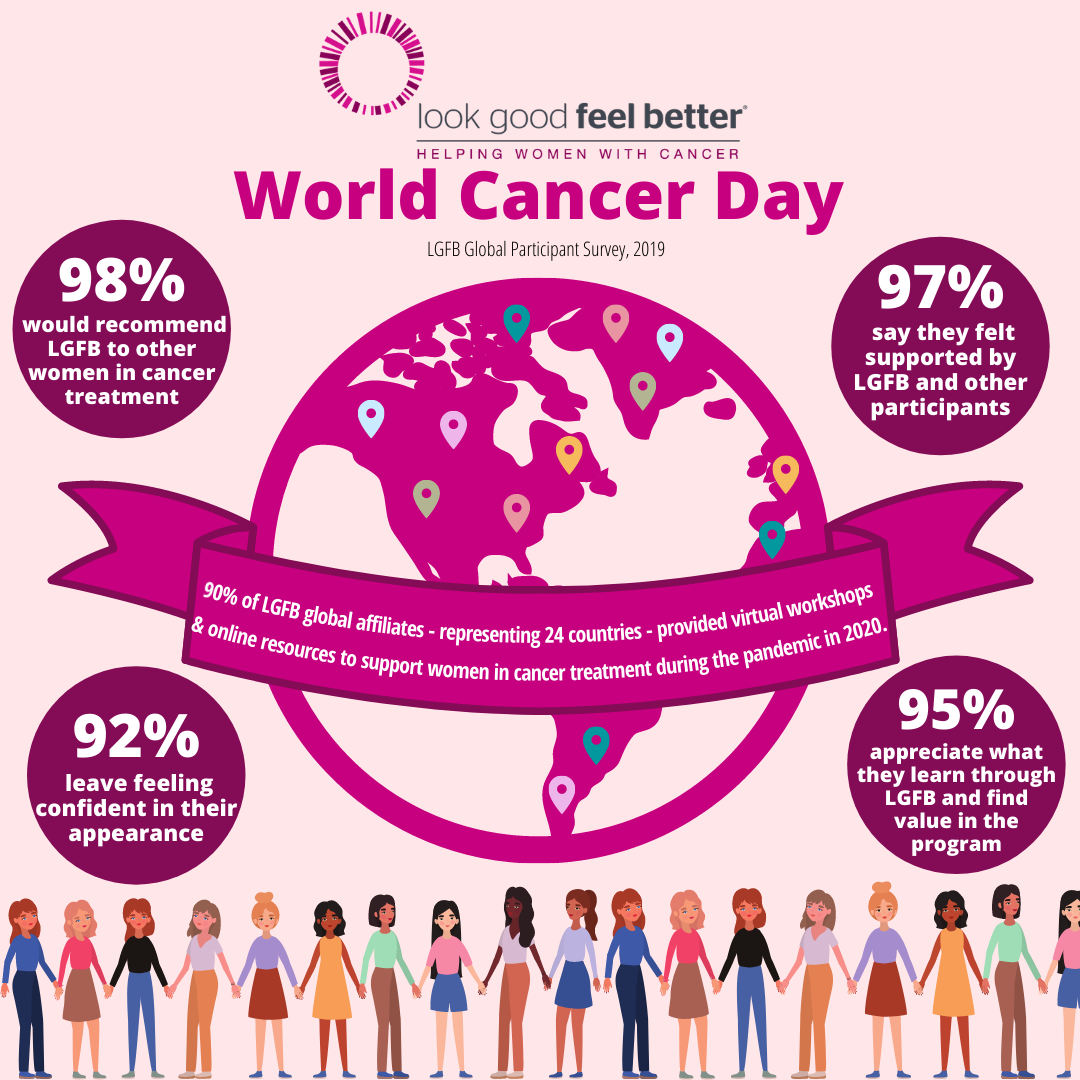}}    & \parbox{3cm}{Anderson Cooper hits Trump with a hard truth about how he'll go down in history} &  \parbox{6cm}{``Seven days after choosing to be the first president ever to incite insurrection against the country and the Constitution he took an oath to defend, the president now has another first to his name,'' Cooper said. ``The first and only president to ever be impeached twice.''}& \parbox{1.5cm}{\centering 0} & \parbox{2cm}{\centering dEFEND : 1 \\ HPFN : 1\\ MCAN : 1\\ CPDM : 1\\ CCD : 0\\ VLP : 0\\Ours : 0}   \\ \cmidrule[0.5pt](l{1pt}r{0pt}){1-5}
        
\raisebox{-0.4\height}{\includegraphics[width=1.5cm]{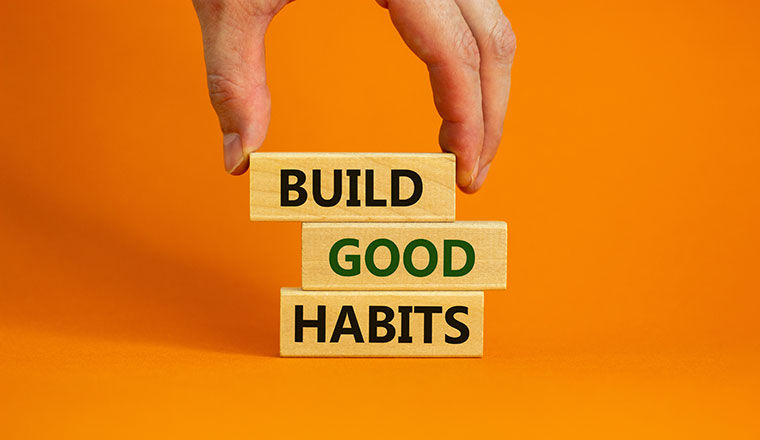}}  & \parbox{3cm}{Bid Rupert Murdoch farewell by reliving his most controversial tweets} & \parbox{6cm}{Rupert Murdoch, one of the top media executives in the world, is stepping down from his perch as CEO of 21st Century Fox…``Independence Day. Immigration is our history and immigration MUST be our future. Multi-ethnicities and equality under law for all.``Making rich poorer won't do much. Giving opportunity too poor to become rich is the only way forward.''} & \parbox{1.5cm}{\centering 1} & \parbox{2cm}{\centering \centering dEFEND : 0 \\ HPFN : 0\\ MCAN : 0\\ CPDM : 0\\ CCD : 1\\ VLP : 0\\Ours : 1}  \\ 
\bottomrule[1.2pt]
\end{tabular}}
\caption{Qualitative analysis of test cases for various methods. 0 indicates non-clickbait, while 1 denotes clickbait.}
\label{tab:case study}
\end{table*}

\begin{table}[h]
 \resizebox{\linewidth}{!}{
   \begin{tabular}{c|ccc}
     \toprule[1.2pt]
     Dataset & \#Samples & \#Clickbait & \#non-Clickbait\\
     \midrule[0.8pt]
     \makecell[c]{WeChatCB}  & 70,785  & 32,418   & 38,367\\
   \bottomrule[1.0pt]
 \end{tabular}
}
\caption{The statistics on the pseudo-labeled samples.}
\label{tab:augstatistic1}
\end{table}

\begin{table}[!h]
\centering
\resizebox{\linewidth}{!}{
\begin{tabular}{c@{\hspace{5pt}}|c@{\hspace{4.8pt}}c@{\hspace{4.8pt}}|c@{\hspace{4.8pt}}c@{\hspace{4.8pt}}|c@{\hspace{4.8pt}}c@{\hspace{4.8pt}}}
\toprule[1.2pt]
\multirow{2}{*}{Datasets}  & \multicolumn{2}{c|}{\cellcolor{gray!15}\textbf{CLDInst}} & \multicolumn{2}{c|}{\cellcolor{gray!15}\textbf{Clickbait17}} & \multicolumn{2}{c}{\cellcolor{gray!15}\textbf{FakeNewsNet}} \\
        & PRE $\downarrow$   & REC $\downarrow$ & PRE $\downarrow$   & REC $\downarrow$ & PRE $\downarrow$   & REC  $\downarrow$  \\ \midrule[0.8pt]
w/o MF   
& -4.59\tiny{$\pm$ 0.14}  & -4.85\tiny{$\pm$ 0.52} & -5.81\tiny{$\pm$ 0.37}   & -6.11\tiny{$\pm$ 0.29} & -5.42\tiny{$\pm$ 0.22}   & -5.79\tiny{$\pm$ 0.04}             \\
w/o EICF 
& \textbf{-5.89}\tiny{$\pm$ 0.45}   & \textbf{-6.16}\tiny{$\pm$ 0.17} & \textbf{-7.79}\tiny{$\pm$ 0.11}    & \textbf{-8.32}\tiny{$\pm$ 0.30} & \textbf{-6.90}\tiny{$\pm$ 0.26}  & \textbf{-7.23}\tiny{$\pm$ 0.19}         \\
w/o ESCF 
& -3.53\tiny{$\pm$ 0.43}   & -3.87\tiny{$\pm$ 0.55} & -4.02\tiny{$\pm$ 0.29}  & -4.52\tiny{$\pm$ 0.06}  & -3.81\tiny{$\pm$ 0.14}  & -3.93\tiny{$\pm$ 0.11}   \\
w/o ENF  
& -3.78\tiny{$\pm$ 0.35}   & -4.06\tiny{$\pm$ 0.12} & -4.62\tiny{$\pm$ 0.51}  & -4.85\tiny{$\pm$ 0.43}  & -4.14\tiny{$\pm$ 0.06}  & -4.49\tiny{$\pm$ 0.40}   \\ \bottomrule[1.0pt]
\end{tabular}}
\caption{Ablation studies. T-test, p-value$<$0.005.}
\label{tab:ablationother1}
\end{table}

\subsection{Study of Data Augmentation Technique}
\label{sec:augmentationdetail}
%考虑到现有标注数据集在规模和多样性方面都不足，我们提出了基于用户反馈生成伪标签样本的数据增强方法。为了验证该方法的有效性，我们从微信收集了3w条无标注帖子并生成相应的伪标签，统计结果如表所示。将这些数据作为训练集学习我们模型和baselines的参数，并在三个数据集的测试集上验证了性能。实验结果如表所示，结合表1的结果可以发现，所有方法在Webis16上都有了提升，且SVM不再优于其他深度学习方法。我们认为这是因为伪标签数据为深度学习模型提供了相当的数据驱动力而导致的。此外，在其余两个数据集上，几乎所有方法都能与原来的性能持平，而VLP甚至有了性能提升。我们认为这是由于伪标签训练集的标题党-非标题党数据更平均导致的

In real applications, training data is usually insufficient or even not. To address this problem, we develop a data augmentation technique which collected pseudo-labeled samples from the historical data in social media, e.g. \textit{WeChat}. We found that although the social behavior-based approach needed extra cold start time, its accuracy was high if sufficient feedback data was accumulated over time. We thus propose to collect the user behavior of these posts from historical data over a period of time, so as to build a batch of pseudo-labeled data. In detail, we first collected hot posts with more than $\mu=100,000$ forward actions per hour. We then designed two heuristic rules to identify clickbait, including those that take less than $\nu=10$ seconds viewing duration; and the number of user likes is 0. The remaining posts were viewed as non-bait. The statistics on dataset are shown in Table~\ref{tab:augstatistic1}. We used the \textit{Google Translate API} to translate these Chinese posts into English, and got a data-augmented dataset \emph{WeChatCB}. To analyze the quality of this dataset, we randomly selected 500 samples for human evaluation. The result showed that approximately $28\%$ of the posts were disguised type. It indicated that \textit{WeChat}, like other popular social platforms, existed serious deceptive issue. 

To verify the effectiveness of this augmentation technique, we simulated a few-shot and zero-shot environment and trained our model on the augmented data. If the quality of this data is equivalent to the human-tagged one, their performance should be comparable. In detail, we retained $10\%$ of the training data from the original dataset and randomly selected data with a size of $90\%\sim190\%$ from the augmented dataset as an additional training set. The selected data included both positive and negative samples. The results on the test sets of \emph{CLDInst}, \emph{Clickbait17}, and \emph{FakeNewsNet}, respectively are shown in Table~\ref{tab:aug_performance}. Moreover, we also showed the results of using only the augmented samples (from a size of $100\%$ to $200\%$), but with no original training data. Based on 10\% human-tagged data and 150\% machine-tagged data, our model obtained better performance than the one based on 100\% original training data (i.e., human-tagged data). When in the zero-shot situation, our model still can obtain good performance based on 180\% machine-tagged data. The outperformance would be larger when feeding more tagged data. We can infer that the quality of machine tagging is acceptable and satisfactory. Our data augmentation technique is useful to alleviate for the shortage problem of human-tagged resources. 

\subsection{Ablation Studies on Other Metrics}\label{sec:ablation_other}
%%%%%%%%%%%%%%%%%%% Ablation Study Table %%%%%%%%%%%%%%%%%%%

As demonstrated in Table~\ref{tab:ablationother1}, we plotted the results in terms of \textit{PRE} and \textit{REC} metrics, respectively.

\begin{table*}[!h]
\centering
\small
\resizebox{\linewidth}{!}{
    \begin{tabular}{c|c@{\hspace{3pt}}c@{\hspace{3.2pt}}c@{\hspace{3.2pt}}c@{\hspace{3.2pt}}|c@{\hspace{3.2pt}}c@{\hspace{3.2pt}}c@{\hspace{3.2pt}}c@{\hspace{3.2pt}}|c@{\hspace{3.2pt}}c@{\hspace{3.2pt}}c@{\hspace{3.2pt}}c@{\hspace{3.2pt}}}
    \toprule[1.2pt]
    \multirow{2}{*}{Data Settings} & \multicolumn{4}{c|}{\cellcolor{gray!15}\textbf{CLDInst}}     & \multicolumn{4}{c|}{\cellcolor{gray!15}\textbf{Clickbait17}}     & \multicolumn{4}{c}{\cellcolor{gray!15}\textbf{FakeNewsNet}}   \\
    & ACC     & PRE     & REC    & F1      & ACC   & PRE     & REC  & F1  & ACC    & PRE   & REC    & F1      \\ \midrule[0.75pt]
    
    \textbf{\footnotesize{H 100\% }}
    & 88.79\tiny{$\pm$ 0.11} & 88.73\tiny{$\pm$ 0.04} & 89.16\tiny{$\pm$ 0.32} & 88.94\tiny{$\pm$ 0.14} & 92.83\tiny{$\pm$ 0.27} & 93.45\tiny{$\pm$ 0.17} & 93.59\tiny{$\pm$ 0.15} & 93.52\tiny{$\pm$ 0.20} & 91.56\tiny{$\pm$ 0.07} & 92.74\tiny{$\pm$ 0.38} & 92.97\tiny{$\pm$ 0.13} & 92.85\tiny{$\pm$ 0.08}  \\
    \midrule[0.75pt]
    
    \textbf{\footnotesize{H 10\%  +  M 90\% } }
    & 83.25\tiny{$\pm$ 0.31} & 85.99\tiny{$\pm$ 0.22} & 84.40\tiny{$\pm$ 0.05} & 85.19\tiny{$\pm$ 0.17} & 87.95\tiny{$\pm$ 0.58} & 90.16\tiny{$\pm$ 0.03} & 90.56\tiny{$\pm$ 0.71} & 90.36\tiny{$\pm$ 0.24} & 86.19\tiny{$\pm$ 0.36} & 89.64\tiny{$\pm$ 0.49} & 88.99\tiny{$\pm$ 0.10} & 89.31\tiny{$\pm$ 0.12}  \\
    
    \textbf{\footnotesize{H 10\%  + M 110\% }}       
    & 85.53\tiny{$\pm$ 0.06} & 87.37\tiny{$\pm$ 0.42} & 86.66\tiny{$\pm$ 0.33} & 87.01\tiny{$\pm$ 0.50} & 90.27\tiny{$\pm$ 0.21} & 91.84\tiny{$\pm$ 0.16} & 91.70\tiny{$\pm$ 0.47} & 91.77\tiny{$\pm$ 0.13} & 88.06\tiny{$\pm$ 0.11} & 91.33\tiny{$\pm$ 0.27} & 90.85\tiny{$\pm$ 0.24} & 91.09\tiny{$\pm$ 0.09}  \\
    
    \textbf{\footnotesize {H 10\%  + M 130\% }}
    & 88.22\tiny{$\pm$ 0.32} & 88.51\tiny{$\pm$ 0.24} & 88.37\tiny{$\pm$ 0.08} & 88.44\tiny{$\pm$ 0.17} & 92.62\tiny{$\pm$ 0.14} & 93.06\tiny{$\pm$ 0.49} & 92.86\tiny{$\pm$ 0.36} & 92.96\tiny{$\pm$ 0.10} & 90.21\tiny{$\pm$ 0.21} & 92.28\tiny{$\pm$ 0.40} & 92.53\tiny{$\pm$ 0.29} & 92.41\tiny{$\pm$ 0.16}  \\
    
    \textbf{\footnotesize{H 10\%  +  M 150\% }}       
    & 89.87\tiny{$\pm$ 0.03} & 89.82\tiny{$\pm$ 0.13} & 89.38\tiny{$\pm$ 0.51} & 89.60\tiny{$\pm$ 0.44} & 93.80\tiny{$\pm$ 0.28} & 93.33\tiny{$\pm$ 0.24} & 93.72\tiny{$\pm$ 0.30} & 93.52\tiny{$\pm$ 0.15} & 92.02\tiny{$\pm$ 0.44} & 92.84\tiny{$\pm$ 0.38} & 93.05\tiny{$\pm$ 0.07} & 92.95\tiny{$\pm$ 0.32}  \\ 
    
    \textbf{\footnotesize{H 10\%  + M 170\%}}          
    & 91.85\tiny{$\pm$ 0.24} & 91.28\tiny{$\pm$ 0.35} & 90.51\tiny{$\pm$ 0.56} & 90.89\tiny{$\pm$ 0.17} & 94.07\tiny{$\pm$ 0.13} & 93.71\tiny{$\pm$ 0.25} & 94.25\tiny{$\pm$ 0.08} & 93.98\tiny{$\pm$ 0.04} & 93.00\tiny{$\pm$ 0.52} & 93.47\tiny{$\pm$ 0.55} & 93.98\tiny{$\pm$ 0.09} & 93.73\tiny{$\pm$ 0.31}  \\

     \textbf{\footnotesize{H 10\%  + M 190\% }}          
    & 92.15\tiny{$\pm$ 0.51} & 92.51\tiny{$\pm$ 0.22} & 92.57\tiny{$\pm$ 0.38} & 92.54\tiny{$\pm$ 0.05} & 94.43\tiny{$\pm$ 0.34} & 94.35\tiny{$\pm$ 0.27} & 94.64\tiny{$\pm$ 0.18} & 94.49\tiny{$\pm$ 0.44} & 94.07\tiny{$\pm$ 0.08} & 94.37\tiny{$\pm$ 0.20} & 94.48\tiny{$\pm$ 0.54} & 94.42\tiny{$\pm$ 0.42}  \\ \midrule[0.75pt]

    \textbf {\footnotesize{M 100\% }}
    & 82.90\tiny{$\pm$ 0.56} & 85.29\tiny{$\pm$ 0.38} & 85.42\tiny{$\pm$ 0.21} & 85.35\tiny{$\pm$ 0.60} & 87.20\tiny{$\pm$ 0.44} & 90.05\tiny{$\pm$ 0.05} & 88.71\tiny{$\pm$ 0.19} & 89.37\tiny{$\pm$ 0.24} & 87.89\tiny{$\pm$ 0.43} & 89.52\tiny{$\pm$ 0.51} & 86.25\tiny{$\pm$ 0.11} & 87.85\tiny{$\pm$ 0.39}  \\

    \textbf {\footnotesize{M 120\% }} 
    & 84.46\tiny{$\pm$ 0.18} & 86.46\tiny{$\pm$ 0.05} & 86.64\tiny{$\pm$ 0.14} & 86.55\tiny{$\pm$ 0.62} & 88.30\tiny{$\pm$ 0.30} & 90.91\tiny{$\pm$ 0.11} & 89.66\tiny{$\pm$ 0.25} & 90.28\tiny{$\pm$ 0.32} & 89.33\tiny{$\pm$ 0.35} & 90.93\tiny{$\pm$ 0.50} & 88.03\tiny{$\pm$ 0.12} & 89.46\tiny{$\pm$ 0.29}  \\

    \textbf {\footnotesize{M 140\% }}
    & 86.22\tiny{$\pm$ 0.24} & 87.41\tiny{$\pm$ 0.30} & 87.58\tiny{$\pm$ 0.58} & 87.49\tiny{$\pm$ 0.16} & 90.26\tiny{$\pm$ 0.29} & 91.89\tiny{$\pm$ 0.08} & 90.62\tiny{$\pm$ 0.03} & 91.26\tiny{$\pm$ 0.29} & 90.82\tiny{$\pm$ 0.51} & 91.51\tiny{$\pm$ 0.40} & 89.15\tiny{$\pm$ 0.05} & 90.32\tiny{$\pm$ 0.14}  \\

    \textbf {\footnotesize{M 160\% }}
    & 87.76\tiny{$\pm$ 0.62} & 87.92\tiny{$\pm$ 0.40} & 88.30\tiny{$\pm$ 0.18} & 88.11\tiny{$\pm$ 0.09} & 91.78\tiny{$\pm$ 0.23} & 92.75\tiny{$\pm$ 0.41} & 91.77\tiny{$\pm$ 0.57} & 92.26\tiny{$\pm$ 0.54} & 91.98\tiny{$\pm$ 0.34} & 92.02\tiny{$\pm$ 0.50} & 91.33\tiny{$\pm$ 0.21} & 91.67\tiny{$\pm$ 0.26}  \\ 

    \textbf {\footnotesize{M 180\% }}
    & 89.14\tiny{$\pm$ 0.43} & 89.20\tiny{$\pm$ 0.16} & 88.85\tiny{$\pm$ 0.22} & 89.03\tiny{$\pm$ 0.24} & 92.54\tiny{$\pm$ 0.17} & 93.36\tiny{$\pm$ 0.62} & 93.91\tiny{$\pm$ 0.07} & 93.63\tiny{$\pm$ 0.35} & 92.46\tiny{$\pm$ 0.26} & 92.47\tiny{$\pm$ 0.06} & 92.30\tiny{$\pm$ 0.45} & 92.38\tiny{$\pm$ 0.10}  \\

    \textbf {\footnotesize{M 200\% }}
    & 90.11\tiny{$\pm$ 0.14} & 90.89\tiny{$\pm$ 0.39} & 90.75\tiny{$\pm$ 0.06} & 90.82\tiny{$\pm$ 0.42} & 94.02\tiny{$\pm$ 0.64} & 93.78\tiny{$\pm$ 0.19} & 93.58\tiny{$\pm$ 0.20} & 93.68\tiny{$\pm$ 0.18} & 93.31\tiny{$\pm$ 0.15} & 93.08\tiny{$\pm$ 0.53} & 93.88\tiny{$\pm$ 0.33} & 93.48\tiny{$\pm$ 0.07}  \\
    \bottomrule[1pt]
    \end{tabular}
    }
\caption{Evaluation of the quality of the augmented data. H and M represent the human-tagged data and machine-augmented data which are used to train our method, respectively.}
\label{tab:aug_performance}
\end{table*}

%%%%%%%%%%%%%%%%%%%% Number of the scenarios %%%%%%%%%%%%%%%%%%%%%

\end{document}